\documentclass{article} 
\usepackage{nips14submit_e,times}
\usepackage{hyperref}
\usepackage{url}
\usepackage{natbib}
\usepackage{amsmath}
\usepackage{amsthm}
\usepackage{amssymb}
\usepackage{graphicx}
\usepackage{xspace}
\usepackage{tabularx}
\usepackage{algorithm}
\usepackage{algorithmic}

\newif\ifarxiv
\arxivtrue

\newcommand{\bmx}[0]{\begin{bmatrix}}
\newcommand{\emx}[0]{\end{bmatrix}}

\nipsfinalcopy 

\title{How Auto-Encoders Could Provide Credit Assignment in Deep Networks via Target Propagation}

\author{
Yoshua Bengio \\
Universit\'{e} de Montr\'{e}al \\ CIFAR Fellow
}

\begin{document}

\maketitle

\begin{abstract}
We propose to exploit {\em reconstruction} as a layer-local
training signal for deep learning. Reconstructions can be propagated in a form of target
propagation playing a role similar to back-propagation but helping to
reduce the reliance on derivatives in order to perform credit
assignment across many levels of possibly strong non-linearities (which is
difficult for back-propagation). A regularized auto-encoder tends produce a
reconstruction that is a more likely version of its input, i.e., a small
move in the direction of higher likelihood. By generalizing gradients,
target propagation may also allow to train deep networks with discrete
hidden units.  If the auto-encoder takes both a representation of input and
target (or of any side information) in input, then its reconstruction of
input representation provides a target towards a representation that is
more likely, conditioned on all the side information. A deep auto-encoder
decoding path generalizes gradient propagation in a learned way that can
could thus handle not just infinitesimal changes but larger, discrete changes,
hopefully allowing credit assignment through a long chain of non-linear
operations.  In addition to each layer being a good
auto-encoder, the encoder also learns to please the upper layers by
transforming the data into a space where it is easier to model by them,
flattening manifolds and disentangling factors.  The motivations and
theoretical justifications for this approach are laid down in this paper,
along with conjectures that will have to be verified either mathematically
or experimentally, including a hypothesis stating that such auto-encoder
mediated target propagation could play in brains the role of credit
assignment through many non-linear, noisy and discrete transformations.
\end{abstract}

\section{Introduction}

Deep learning is an aspect of machine learning that regards the question
of learning multiple levels of representation, associated with different
levels of abstraction~\citep{Bengio-2009-book}. 
These representations are distributed~\citep{Hinton89b}, meaning that
at each level there are many variables or features, which together can take a very
large number of configurations. Deep representations can be learned
in a purely unsupervised way (sometimes with a generative procedure
associated with the model), in a purely supervised way (e.g., a deep
feedforward network), or in a semi-supervised way, e.g., with unsupervised
pre-training~\citep{Hinton06,Bengio-nips-2006-small,ranzato-07}. It
is possible to build deep representations that capture the relationships
between multiple modalities~\citep{Srivastava+Salakhutdinov-2014}
and to model sequential structure through recursive application~\citep{tr-bottou-2011}
of learned representation-to-representation transformations, e.g.
with recurrent neural networks~\citep{Sutskever-thesis2012} or 
recursive neural networks~\citep{Socher-2011-small}.

Deep learning methods have essentially relied on three approaches to
propagate training signals across the different levels of representation:
 \begin{enumerate}
\item Greedy layer-wise pre-training~\citep{Hinton06,Bengio-nips-2006-small,ranzato-07}: 
the lower levels are trained
without being influenced by the upper levels, each only trying to find
a better representation of the data as they see it from the output
of the previous level of representation. Although this approach
has been found very useful as an initialization for the second
or third approach, its potential disadvantage
is that by itself it does not provide a global coordination between
the different levels.
\item Back-propagated gradients: the features must be continuous and
  gradients with respect to lower layers (through many non-linearities)
  appear less able to correctly train
  them, especially when considering deeper networks trying to
  learn more abstract concepts obtained from the composition of simpler 
  ones~\citep{Bengio-2009-book,Gulcehre+Bengio-ICLR2013}.  This is
  especially true of very deep or recurrent
  nets~\citep{Hochreiter91-small,Bengio-trnn93-small,Hochreiter+Schmidhuber-1997,Pascanu+al-ICML2013-small},
  involving the composition of many non-linear operations. Back-propagation
  and gradient-based optimization can be used in a supervised setting
  (e.g., for classification or regression) or an unsupervised setting, e.g.,
  for training shallow~\citep{VincentPLarochelleH2008-small} or
  deep~\citep{martens2010hessian-small} auto-encoders and deep generative
  models~\citep{Goodfellow-et-al-ARXIV2014}.
\item Monte-Carlo Markov chains (MCMC): the stochastic relaxation performed
in models such as Markov random fields and Boltzmann machines (including
the Restricted Boltzmann Machines) propagates information about the
observed variables into the parameters governing the behavior of
latent variables (hidden units), so as to make the sufficient statistics of configurations
generated by the model as close as possible as to those obtained
when the observed visible units are clamped. Unfortunately, with some form 
of MCMC in the inner loop of training,  mixing difficulties may limit
our ability to learn models that assign probability in a sharp way,
near manifolds (modes) that are far from each other~\citep{Bengio-et-al-ICML2013,Bengio-arxiv-2013}.
This becomes especially troublesome when one tries to learn models of
complex distributions involving a manifold structure in high dimension,
as in common AI tasks (involving images, text, sound, etc.).
\end{enumerate}

There is a fourth and insufficiently explored approach, to which the
proposal discussed here belongs, based on {\em target
  propagation}~\citep{LeCun-dsbo86}. Back-propagation and target
propagation are identical when the target is viewed as an infinitesimal
direction of change and the gradient can be computed
analytically~\citep{Lecun-these87}. However, target propagation
can also be potentially applied to the case where the representation
involves discrete values~\citep{LeCun-dsbo86}. Target propagation
was previously also proposed~\citep{Bengio-trnn93-small} 
as a way to defeat some of the optimization
difficulties with training recurrent neural networks, due to long-term
dependencies in sequential data.
This target propagation viewpoint is
also related to optimization
approaches where the representation values (i.e., hidden unit
activations) are free variables that can be optimized
over~\citep{Carreira-Perpinan-and-Wang-AISTATS2014}.
There are also iterative forms of target propagation
such as in the Almeida-Pineda recurrent networks~\citep{Almeida87,Pineda87}
and in the generalized recirculation algorithm~\citep{OReilly-1996}.
In this paper, we propose to use auto-encoders (or conditional auto-encoders,
whose ``code'' can depend on side information as well) 
to provide learned reconstructions that can be used as targets
for intermediate layers. In a sense, we are proposing to
{\em learn the back-propagation computation}. In doing so,
we generalize back-propagation (for example allowing one to
handle discrete-valued elements of the representation) and
make it more biologically plausible as a mechanism for
brains to perform credit assignment through many levels of
a deep computation, including a temporally recurrent one.
A recent paper proposes the reweighted wake-sleep algorithm,
that learns a deep generative model and can
also handle discrete latent variables~\citep{Bornschein+Bengio-arxiv2014-small},
while being based on a generalization of the wake-sleep algorithm~\citep{Hinton95}.
The proposed target propagation indeed has some similarity to both the original wake-sleep
algorithm and the reweighted wake-sleep algorithm. In fact, the two criteria
become identical when the encoder is deterministic.

This paper starts by what is probably the simplest and most natural context
for the idea of using reconstruction for target propagation: that of
generative models in which each layer is trained as an auto-encoder, and at
any level, the representation $h$ has a distribution prior $P(H)$ which is
implicitly captured by the upper levels. The mathematical framework we propose
is based on recent work on variational auto-encoders~\citep{Kingma+Welling-ICLR2014,Rezende-et-al-arxiv2014,Gregor-et-al-ICML2014},
i.e., a lower bound on the log-likelihood, but can also be interpreted as trying
to match the joint distribution of latent and observed variables under the generative
model and under the composition of the data generating distribution with
a ``recognition network''~\citep{Hinton95} that predicts latent variables given inputs.
The main idea is that upper auto-encoders provide by their reconstruction
a proposed change which indicates the direction of higher prior probability,
and can be used as a proxy for the gradient towards making the lower
levels produce outputs that are more probable under the implicit model
of the upper levels.

This paper then extends this framework
to the classical supervised setting in which both an input variable $x$ and
a target or output variable $y$ are involved. In this case, when $y$ is
observed, it constrains the reconstruction of a layer $h$ that was initially
computed based only on $x$, i.e., the
reconstruction provides a representation value that is compatible with
both $x$ and $y$. If $h(x)$ is the encoding of $x$, then the reconstruction
of $h$ by an auto-encoder that also sees $y$ can be used to estimate
$\frac{\partial \log P(y , h(x))}{\partial h(x)}$ (the joint likelihood criterion) or
$\frac{\partial \log P(y | h(x))}{\partial h(x)}$ (the discriminant criterion).
This approach can be naturally generalized to multi-modal
data, where instead of $x$ and $y$, one can think of different sensory modalities $x^{(i)}$.
Again, the reconstruction of a representation $h^{(i)}$ which initially only
depends on $x^{(i)}$ is made to depend on both $x^{(i)}$ and the observations
made with other modalities $x^{(j)}$.
By generalizing the form of the unfolded graph of computations,
these ideas can be extended to the very interesting context
of recurrent networks, in which we have a sequence of $x_t$'s, and
the reconstruction of a representation $h_t$ of the past sequence
incorporates the constraints induced by future observations.
Finally, this paper discuss the potential of this framework to
provide a biologically plausible credit assignment mechanism
that would replace and not require back-propagation.

\section{Stacked Auto-Encoders as Generative Models}

\subsection{Preliminaries}

\subsubsection{Notation}

Denote $h_l$ the layer-$l$ representation, which we generally
think of as a random variable, and \mbox{$h=(h_1, h_2, \ldots)$} all
the layers of a generative model. Denote $p(x,h)$ the joint
distribution over $x$ and $h$, structured as a directed graphical model 
\mbox{$h_L \Rightarrow \ldots h_2 \Rightarrow h_1 \Rightarrow x$}
with a {\em chain structure}:

\begin{equation}
\label{eq:chain}
 P(x,h) = P(x|h_1)P(h_1|h_2) \ldots P(h_{L-1}|h_L)P(h_L).
\end{equation}
With $x=h_0$, one can view each $P(H_l | h_{l+1})$ as one
layer of a generative network, or {\em decoder} network,
that maps top-level representations into low-level samples.

This graphical structure is the same as in sigmoidal belief
networks~\citep{Neal92}, but just like for Helmholtz 
machines~\citep{Dayan-et-al-1995}, we will also consider
a {\em recognition} network, or {\em encoder}, or approximate inference
network, which computes in the reverse direction,
starting with the unknown data generating distribution, which we
denote $Q(X)$:
\begin{equation}
 Q(x,h) = Q(h_L|h_{L-1})\ldots Q(h_2|h_1) Q(h_1 | x) Q(x).
\end{equation}
Accordingly, we define $P(H_l)$ and $Q(H_l)$ as the 
marginals respectively associated with the joints $P$ and $Q$
over all the variables.

\subsubsection{Motivating Deterministic Decoders}

Most deep generative models proposed in the past, such as deep Belief
networks~\citep{Hinton06}, Helmholtz machines~\citep{Dayan-et-al-1995} and
deep Boltzmann machines~\citep{Salakhutdinov2009-small}, have the property
that lots of ``noise'' gets injected at every level of the multilayer
generative model. At the lowest level ($P(H|h_1)$), it typically means that the
generated $x$ is the ``addition'' of an expected $E[X|h_1]$ and some iid
``noise''. This iid noise shows up in generated samples as high-frequency
spatial (for images) or temporal (for acoustics) noise which is not at all
similar to what is observed in the training data. It necessarily moves away
from any low-dimensional manifold near which the distribution concentrates,
ruining the possibility of being able to capture such a sharp distribution. 

A similar remark can be made regarding the noise injected when sampling
$h_1$ given $h_2$, because the mapping from $h_1$ to $x$ is generally
simple (e.g. affine + simple saturating non-linearity). 
The effect of such noise is especially striking when $h_1$ is a
vector of stochastic binary units: when $h_1$ is sampled from $P(H_1|h_2)$,
many independent coin flips are generated to choose the different values
$h_{1i}$.  These independent sources of noise must then somehow be transformed into
a well-formed $x$ through a simple mapping (such as the affine transformation
composed with sigmoidal non-linearity typically used in such models). This
could only happen if both $P(X|h_1)$ and $P(H_1|h_2)$ are almost
deterministic (i.e. no coin flips), or if the dimension of $h_1$ is so large that the
independent noise sources cancel each other. Note that if $P(X|h_1)$ and
$P(H_1|h_2)$ are nearly deterministic (i.e., nearly diracs), then the learning procedures
typically proposed for such models break down. For example, when the
weights of a Boltzmann machine become large, MCMCs do not mix and training
stalls. 

Another, very different and very interesting view of a deep generative model is offered in the
recent work on adversarial generative
networks~\citep{Goodfellow-et-al-ARXIV2014}. In that case, randomness
can be viewed as injected at the top level (possibly at lower layers too,
but that is a choice of the designer), while the intermediate levels
are considered to be purely deterministic and continuous transformations.
What could make learning in such networks difficult is that one still
has to assign credit (by backprop) to all the layers of a deep net.
Another less well understood potential issue is that for each update step of the
generator network, a discriminator network must be sufficiently
re-optimized to continuously track the decision surface between
examples generated by the model and training examples. Finally, another
potential issue (which can be seen as part of the optimization problem)
is that if the generator concentrates on some modes of the data distribution,
there is very little pressure to make it generate samples for the other modes
(only for the rare generated samples that approach these other modes does the generator
get a signal). However, a great innovation of the adversarial network
is that it opens the door to deep generative models whose layers are trained in
a globally coordinated way and in which the noise is injected
at the top level, which we consider here to be very important features.

In any case, the basic idea in many of these models is that although the
top-level prior $P(H_L)$ is going to be simple, e.g., factorial or a single RBM,
the lower levels gradually transform $P(H_L)$ into more complex
distributions $P(H_l)$, ending in $P(X)$. For example, with the manifold
learning view, we can imagine $P(H_L)$ as essentially uniform on
one or several manifolds where the distribution concentrates, with $h_L$
representing a coordinate system for the data in an abstract space
where all the variables are independent. That flat manifold
is then distorted (and possibly broken down into separate sub-manifolds
for different classes) in complicated non-linear ways. In this view, the
job of the generative network is really just to transform and distort
the space such that a simple distribution gets mapped into a good
approximation of the data generating distribution. Similarly, the
job of the encoder networks (the $Q(h|x)$) is to map a complicated
distribution into a simpler one, layer by layer, to map a highly
curved manifold into a flat one. Under that
manifold-learning perspective, we want most of the ``noise'' injected
when sampling from the model to be injected high in the hierarchy. 
That ``noise'' represents the high-level choices about the content of the generated $x$.
For example, the top-level factors in a deep net modeling images
of a single object might include not just the object category
but all of its attributes, geometrical pose parameters, lighting
effects, etc. We know from the physics of image synthesis that
the mapping from such high-level variables to pixels is highly
non-linear, which means that these factors should be chosen
(i.e. ``sampled'') high up in the deep network.

\begin{figure}[ht]
\begin{center}
\ifarxiv
\centerline{\includegraphics[width=0.5\columnwidth]{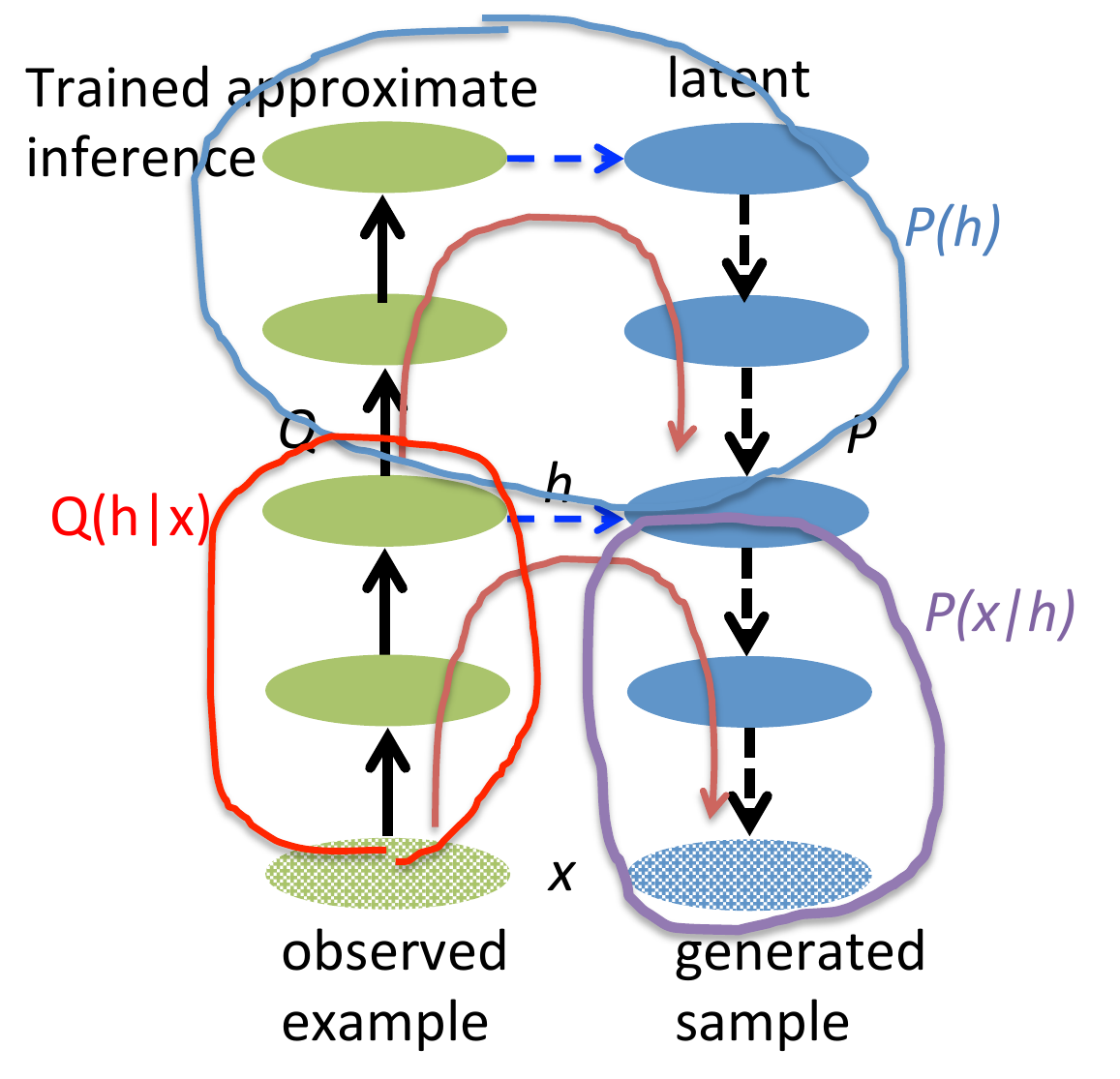}}
\else
\centerline{\includegraphics[width=0.5\columnwidth]{figures/criterion-decomposition.pdf}}
\fi
\caption{For any intermediate layer $h$, we consider as generative model
the combination of the top-level prior $P(h)$ (e.g. represented by a deep
denoising auto-encoder), a decoder $P(x|h)$, and an encoder $Q(h|x)$.
The training criterion requires the encoder and decoder to reconstruct $x$ well
(lower auto-encoding loop),
while $P(h)$ models well the transformation of data $x$ into $h$ through $Q(h|x)$,
while the entropy of $Q(h|x)$ is increased where possible. It means that
the top-level auto-encoder (upper auto-encoding loop) does a good job of modeling $h\sim Q(H|x)$
with data $x$, and that the encoder produces samples that are easy to model by $P(h)$.
}
\label{fig:criterion-decomposition}
\end{center}
\end{figure} 

\subsection{A Generative Stack of Auto-Encoders}

\subsubsection{Matching Recognition and Generative Networks}

We propose here that an appropriate objective for training
a deep encoder/decoder pair as introduced above is that the joint distribution
over $h$ and $x$ generated by $P$ matches the joint distribution
generated by $Q$. 
Later, we will see that this can be reduced
to having the marginal distributions $P(H_i)$ and $Q(H_i)$ match
each other in some sense, when the layer-wise auto-encoder pairs are good in terms
of minimizing reconstruction error.
The main actors of such models, $P(h)$, $Q(h|x)$ and $P(x|h$) are illustrated 
in Figure~\ref{fig:criterion-decomposition} along with the associated layers
in a deep generative decoder associated with a deep approximate inference encoder.

The motivation is straightforward: if $P(X,H)$ matches well $Q(X,H)$,
then it necessarily means that its marginals also match well, i.e., the
generative distribution $P(X)$ matches well the data generating
distribution $Q(X)$. This criterion is thus an alternative to
maximum likelihood (which tries to directly match
$P(X)$ to $Q(X)$), with the potential advantage of 
avoiding the associated intractabilities which arise
when latent variables are introduced.

Mathematically, this objective of matching $P(X,H)$ and $Q(X,H)$ can be
embodied by the KL-divergence between $Q$ and $P$, taking $Q$ as the
reference, since we want to make sure that $P$ puts probability mass
everywhere that $Q$ does (and especially where $Q(X)$ does).

The criterion $KL(Q||P)$ can be decomposed in order to
better understand it:
\begin{align}
\label{eq:KL}
 KL(Q||P) =& E_{(x,h)\sim Q(X,H)} \log \frac{Q(x) Q(h|x)}{P(x|h) P(h)} \nonumber \\
          =& - H(Q) - E_{x\sim Q(X)} E_{h\sim Q(H|x)} \log P(x|h) - E_{x\sim Q(X)} E_{h \sim Q(H|x)} \log P(h)
\end{align}
We distinguish three terms:
\begin{enumerate}
\item The {\bf entropy} of the joint distribution of $h$ and $x$ under $Q$.
Because $Q(X)$ is fixed, this turns out to be equivalent to the
average entropy of $Q(H|x)$, for $x\sim Q(X)$:
\begin{align}
   H(Q) =& - E_{x\sim Q(X)} E_{h \sim Q(H|x)} \left(\log Q(x) + \log Q(h|x) \right) \nonumber \\
        =& - E_{x\sim Q(X)} \left( \log Q(x) + E_{h \sim Q(H|x)} \log Q(h|x) \right) \nonumber \\
        =& H(Q(X)) + E_{x\sim Q(X)} H(Q(H|x))
\end{align}
where $H(P(A|b))$ is the entropy of the conditional distribution of $A$, given $B=b$.
This allows us to rewrite the KL criterion as follows:
\begin{align}
\label{eq:KL2}
 KL(Q||P) =& - H(Q(X)) - E_{x\sim Q(X)} H(Q(H|x)) - \nonumber \\
 &  E_{(x,h)\sim Q(X,H)} \log P(x|h) - E_{h\sim Q(H)}  \log P(h)
\end{align}
\item The {\bf match of observed $Q(H)$ to the generative model $P(H)$}, measured by the
log-likelihood of the samples $h \sim Q(H|x), x \sim Q(X)$ according to the prior $P(H)$.
Since both $P(H)$ and $Q(H|x)$ are free, we see that $P(H)$ will try to match the samples
coming from the encoder, but {\em also} that $Q(H|x)$ will try to put $h$ in places
where $P(H)$ is high.
\item The {\bf reconstruction log-likelihood} $\log P(x | h)$, when $h \sim Q(H|x)$. This is the
{\em traditional criterion used in auto-encoders.}
\end{enumerate}

Note that this criterion is equivalent to the training criterion proposed
for the variational auto-encoder~\citep{Kingma+Welling-ICLR2014,Rezende-et-al-arxiv2014},
i.e., it can also be justified as a variational bound on the log-likelihood
$E_{x \sim Q(X)} \log P(x)$, where the bound is tight when $Q(h|x)=P(h|x)$, and
the bound arises simply out of Jensen's inequality or explicitly pulling
out $KL(Q(H|x)||P(H|x))$:
\begin{align}
- E_{x \sim Q(X)} \log P(x) =& - E_{x \sim Q(X)} E_{h \sim Q(H|x)} \log P(x) \nonumber \\
  =& - E_{x \sim Q(X)} E_{h \sim Q(H|x)} \log \frac{P(x,h) Q(h|x)}{P(h|x)Q(h|x)} \nonumber \\
  =& - E_{x \sim Q(X)} E_{h \sim Q(H|x)} \log\frac{Q(h|x)}{P(h|x)}
        + \log \frac{P(x,h)}{Q(h|x)} \nonumber \\
  = & E_{x \sim Q(X)} - KL(Q(H|x)||P(H|x)) - E_{x \sim Q(X)} E_{h \sim Q(H|x)} \log \frac{Q(h|x)}{P(x,h)} \nonumber \\
  \leq & - E_{x \sim Q(X)} E_{h \sim Q(H|x)} \log \frac{Q(h|x)}{P(x,h)} \nonumber \\
  = & KL(Q||P) + H(Q(X))
\end{align}
where the last line comes from inspection of the previous line compared
with Eq.~\ref{eq:KL2}, i.e., we have the (fixed) entropy of the data
generating distribution plus the overall $KL(Q||P)$ that we considered here
as a training criterion.  In the penultimate line we recognize the
variational auto-encoder training criterion
\citep{Kingma+Welling-ICLR2014,Rezende-et-al-arxiv2014}.
Note that this is also equivalent to the criterion used in the 
reweighted wake-sleep algorithm~\citep{Bornschein+Bengio-arxiv2014-small}
when $Q(H|x)$ is exactly deterministic: in that case all the samples of $h \sim Q(H|x)$ are the same
and the normalized importance weights are just equal to 1, and the criterion is exactly
the same as proposed here.

Clearly, as $KL(Q||P)$ is minimized, so is the variational bound that relates $KL(Q||P)$
to the negative log-likelihood becoming tighter, since reducing the discrepency between
the joint, $KL(Q(X,H)||P(X,H))$ should indirectly reduce the discrepency between
the conditionals, $KL(Q(H|X)||P(H|X))$. This is also one of the basic motivations for
all variational approach to training models with latent variables.

\subsubsection{Deterministic Encoders Until the Last Level?}
\label{sec:deterministic}

If $Q(H|x)$ has significant entropy, i.e., we consider $h \sim Q(H|x)$
as the result of a deterministic computation combined with some noise,
then it might be difficult for the decoder $P(X|h)$ to get rid of this noise.
In fact, if the added noise hides any information about $x$, i.e., if
we cannot recover $x$ perfectly given $h$, then some entropy will be
forced upon $P(X|h)$. Note that good reconstruction only needs to occur
for $x \sim Q(X)$, the training data. Whereas adding noise is in general
a non-invertible process, if we know that $x$ lives in a smaller set
of configurations (e.g., a manifold), then one can recover $x$ from $h$
so long as no two probable $x$ (under $Q(x)$) are mapped to the same $h$.
For example, if the noise is added only
orthogonally to the manifold (when projected in $h$-space), 
then no information about $x$ is lost, and
the decoder only needs to be smart enough to contract in those directions
orthogonal to the manifold. 
To avoid losing information by injecting the noise, it would have to be added 
``in the right directions'', i.e., orthogonal to the manifold near which $Q(x)$
concentrates.
But it is at the top level that the model has the best ``view'' of the
manifold, where it has been best flattened. Adding noise anywhere else
would be more likely to hurt, making $P(X|h)$ learn a necessarily more
blurry distribution. However, adding noise only at the top level (i.e.,
with $Q(H_L|h_{L-1})$ being stochastic while $Q(H_l|h_{l-1})$ is deterministic
for $l<L$) might
be beneficiary to make sure that the decoder learns to {\em contract in the
noise directions} (eliminate the noise) in case $P(H_L)$ does not
perfectly match $Q(H_L)$ at the top level. Otherwise, when we sample from $P(H_L)$,
if the samples do not correspond to $h_L$'s obtained from $Q(H_L|x)$, the decoder
will be faced with unusual configurations on which it could generalize in
poor ways: we need to train the decoder to ``clean-up'' the ``noise'' due
to the mismatch between $P(H_L)$ and $Q(H_L)$.

One could also consider the possibility that
information about $x$ is lost but that $P(X|h)$ could model the remaining
uncertainty by having a localized very concentrated output distribution
(for example, the ``noise'' added when sampling from $P(X|h)$ only
translates, rotates, etc., moves that stay on the data manifold). 
However, this is is not compatible with the kind of unimodal and
factorial reconstruction distribution that is typically considered.
The only way to avoid that would
be to make $P(X|h)$ a highly multi-modal complicated
distribution (and this is the route that \citet{Sherjil-et-al-arxiv2014}
have taken). 
This is the kind of complexity we would like to avoid, precisely
by mapping the data to the higher levels of representation where manifolds
are flat and easy to model.
Therefore, if we want to keep $P(X|h)$
unimodal (with a simple partition function), it is hypothesized
here that we are better
off forcing some (or even all) outputs of the encoder (which will carry
the ``signal'') to be deterministic functions of $x$ and allowing 
significant noise to be injected only in the top level.

Consider $Q(H|x)$ a Gaussian
with a mean $\tilde{f}(x)$ and a diagonal variance $\sigma(x)$ (with one value for
each component of $h$). Even though the gradient of the entropy
term with respect to $\tilde{f}(x)$ vanishes 
in expectation (as discussed above), the stochastic
gradient of the entropy term can be arbitrarily large in magnitude
as $\sigma_i(x)$ approaches 0 (for any $i$). More generally,
the same problem will occur if the covariance matrix is not diagonal, so long
as some eigenvector of the covariance matrix goes to 0.
For example, in the scalar case, the stochastic gradient with respect
to $\tilde{f}$ (for a given sample of $h \sim Q(H|x)$) would be proportional
to $\frac{\tilde{f}(x)-h}{\sigma^2(x)}$, which goes to infinity
when $\sigma(x)$ becomes small, {\em even though in average
it could be 0} (as is the case of the gradient due to the
entropy term). However, the good news that this gradient may
be integrated out over $Q(H|x)$ in the Gaussian case~\citep{Kingma+Welling-ICLR2014}:
the entropy term, $\log \sigma(x)$, really only depends
on the amount of injected noise, and that term would push the noise to be as large as possible
while not hurting reconstruction too much, i.e., $\sigma_i(x)$
would be large if $h_i$ is not used to reconstruct $x$.
In general (for other forms of prior), the gradient due to maximizing the entropy should
be considered very carefully because for most problems of interest there should be directions in $h$-space
where the projected data concentrate, yielding ``infinite'' probability in some
directions compared to others.

\subsubsection{A Criterion for Each Layer}

Let us consider each layer $h_l$ of the chain structure
(Eq.~\ref{eq:chain}) and apply the above $KL(Q||P)$
decomposition as if we were only considering $h_l$ as
the latent variable. The objective of this approach
is to provide a training signal for each layer $h_l$
and for the parameters of the associated encoders and decoders
(the ones encoding into $h_l$ and reconstructing
lower layers and the ones encoding and reconstructing $h_l$ from above).
We will consider the upper layers
$h_{l+1}$, $h_{l+2}$, etc. as implicitly providing a prior $P(H_l)$
since they will be trained to implicitly model $h_l \sim Q(H_l)$
through the samples $h_l$ that they see as training data.
Hence the criterion for layer $h_l$ is:
\begin{align}
\label{eq:Ql}
 KL(Q^l||P) =& - H(Q(X)) - E_{x\sim Q(X)} H(Q(H_l|x)) - E_{(x,h_l)\sim Q(X,H_l)} \left[ \log P(x|h_l) + \log P(h_l) \right].
\end{align}
Let us consider each of these terms in turn.
\begin{enumerate}
\item We ignore the first term because it is fixed.
\item The second term is the average entropy of the conditional distribution $Q(H_l|x)$.
If we choose $Q(H_l|x)$ to be noise-free, i.e., the deep encoder $\tilde{f}_l$ that maps
$x$ to $h_l$ is not stochastic and it produces 
\begin{equation}
\label{eq:encoder-output}
  h_l = \tilde{f}_l(x)
\end{equation}
with probability 1, then the entropy of $Q(H_l|x)$ is zero and cannot change,
by this design choice. However, as discussed above, it might be a good idea
to allow the top-level $Q(H_L|h_{L-1})$ to have entropy.
\item The third term is a reconstruction negative log-likelihood. If we
assume that $P(X|h_l)$ is nearly deterministic, then it is captured
by a decoder function $\tilde{g}_l$ that maps $h_l$ to $x$. Then the reconstruction
log-likelihood is minimized
totally if $\tilde{f}_l$ and $\tilde{g}_l$ form what 
we call a {\em perfect auto-encoder relative to $Q(X)$}. More precisely, if $h_l$ is
computed deterministically as in Eq.~\ref{eq:encoder-output}, with $x\sim Q(X)$,
then the third term is totally minimized so long as 
\begin{equation}
  \tilde{g}_l(\tilde{f}_l(x))=x
\end{equation}
when $x\sim Q(X)$. Note that the auto-encoder $\tilde{g}_l \circ \tilde{f}_l$ 
will be a perfect auto-encoder if each of the layer-wise auto-encoders below level $l$
is also a perfect auto-encoder. Let us denote each layer-wise auto-encoder pair
by the encoder $f_l$ and the decoder $g_l$, and accordingly define
\begin{equation}
  \tilde{f}_l(x) = f_l(f_{l-1}(\ldots f_2(f_1(x))))
\end{equation}
and
\begin{equation}
  \tilde{g}_l(h_l) = g_1(g_2(\ldots g_{l-1}(g_l(h_l)))).
\end{equation}
Then we have that
\begin{equation}
 g_i(f_i(h_{i-1}))=h_{i-1},\;\; \forall h_{i-1}\sim Q(H_{i-1}),\;\forall i\;\;\;\; \Rightarrow \;\;\;\;
  \tilde{g}_i(\tilde{f}_i(x))=x, \;\; \forall x \sim Q(X), \; \forall i
\end{equation}
Hence it is enough, to cancel the third term, 
to make sure that each layer $l$ is a perfect auto-encoder
for samples from its ``input'' $h_l \sim Q(H_l)$.
Note that there is an infinite number of invertible mappings $f$ and $g$,
so up to now we have not specified something really ``interesting'' for
the system to learn, besides this ability to auto-encode data at each
layer.
\item The last term is the most interesting one, i.e., matching the marginals
$P(H_i)$ and $Q(H_i)$. At any given layer $l$,
it is doing two things, if we consider the pressure on $Q$ and the pressure
on $P$ separately:
 \begin{enumerate}
 \item This marginal matching term is asking the upper layers to learn a prior $P(H_l)$ that gives high
 probability to the samples $h_l \sim Q(H_l)$. This just means that the
 ``input data'' $h_l \sim Q(H_l)$ seen by the upper layers must be
 well modeled by them. For example, the upper layers prior could
 be represented by a deep denoising auto-encoder trained with $h_l \sim Q(H_l)$ as data.
 In that case we want $h_l$ to be the target for updating the reconstruction of the auto-encoder.
 The input of that auto-encoder could be $h_l$ or a corrupted version of it.
 This is similar to what happens when we train a stack of auto-encoders or a stack of RBMs
 in the usual unsupervised pre-training way, i.e., where the data is propagated upwards
 by the lower layers and used as training data for the upper 
 layers models~\citep{Hinton06,Bengio-nips-2006-small}.
 \item The marginal matching term is also asking the lower layers to choose an invertible encoding $\tilde{f}_l$
 such that the transformation of $Q(X)$ into $Q(H_l)$ yields samples that have
 a high probability under the prior $P(H_l)$. In other words, it is {\em asking the
 lower layers to transform the data distribution (which may be very complicated,
 with many twists and turns) into an easy to model distribution (one that
 the upper layers can capture)}. Note however how the pressure on $Q(H_l)$
 (i.e., on $\tilde{f}_l$) and on $P(H_l)$ are asymmetric. Whereas $P(H_l)$
 is simply trained to model the ``data'' $h_l\sim Q(H_l)$ that it sees, 
 $\tilde{f}_l$ is pressured into producing samples more towards the modes
 of $P(H_l)$, thus tending to make $Q(H_l)$ more concentrated and $\tilde{f}_l$
 contractive. Overall there are thus two forces at play on $\tilde{f}_l$:
to minimize the input space reconstruction error, it would like to spread out
all of the $x$ training examples as uniformly as possible across $h$-space.
But an opposing force is at play, making $Q(H_l)$ concentrate in a few
smaller regions (the modes or manifolds of $P(H_l)$). If it weren't for the reconstruction
error, $\tilde{f}_l$ would just map every training example to one or
a few modes of $P(H_l)$ and $P(H_l)$ would just become highly concentrated
on those modes (maximal contraction maps all inputs to the same point). 
But that would make reconstruction error of the input very high.
So the compromise that seems natural is that $\tilde{f}_l$ contracts
(towards modes of $P(H_l)$) in the local directions that it does not need to represent
because they do not correspond to variations present in the data (i.e., 
it contracts in the directions orthogonal to the manifolds near which $Q(X)$
concentrates). However, it yields a more uniform distribution in the directions
of variation of the data, i.e., on the manifolds. Interestingly, we
can view the pressure from $P(H_l)$ onto the encoder as a {\em regularizer}
that prevents the auto-encoder from just learning a general-purpose
identity function (invertible everywhere). Instead, it is forced to
become invertible only where $Q(X)$ is non-negligible.

 Below we discuss how a training signal
 for the lower-level encoder $\tilde{f}_l$ could be provided by the upper auto-encoder
 that captures $P(H_l)$. 
 \end{enumerate}
\end{enumerate}

Note that if we include $h_0=x$ as one of the layers on which
the above criterion is applied, we see that this is just
a proxy for maximum likelihood: at level 0, the only term that remains
``trainable'' is $E_{x\sim Q(X)} \log P(X)$. If $P(X)$ is estimated
by a deep regularized auto-encoder, then this proxy is the usual training
criterion for a regularized auto-encoder. One reason why
we believe that the criteria for the other layers help is that they provide
a training signal for every intermediate layers of the deep auto-encoder,
thus hopefully making the optimization easier. It also justifies
the top-down DBN-like directed generative procedure associated with Eq.~\ref{eq:chain},
which is not obviously applicable to an arbitrary deep auto-encoder,
as well as the MAP inference procedure discussed in Section~\ref{sec:MAP}.

\subsubsection{How to Estimate a Target}

In the last step of the above decomposition, we are required to
specify a change of $\tilde{f}_l$ such as to move $h_l=\tilde{f}_l(x)$
towards a nearby value $\hat{h}_l$ that has a higher probability
under $P(H_l)$ (or return $\hat{h}_l=h_l$ if $h_l$ is already a mode).

Fortunately, we can take advantage of a previously proven theoretical
result~\citep{Alain+Bengio-ICLR2013}, which has been shown in the case where $h_l$ is continuous
and the training criterion for the auto-encoder is simply squared error.
If a very small quantity of noise is injected in the auto-encoder,
then the difference between the reconstruction $\hat{h}_l$ and the
(uncorrupted) input $h_l$ of the auto-encoder is 
proportional\footnote{the proportionality constant is the variance of
the noise injected during training}
to an estimator of $\frac{\partial \log Q(h_l)}{\partial h_l}$,
where $Q(H_l)$ here denotes the ``true'' distribution of the 
data seen by the auto-encoder (i.e. samples from $Q(H_l))$.
We can therefore consider $\hat{h}-h$ as a proxy for the
direction of the gradient $\frac{\partial \log P(h_l)}{\partial h_l}$,
where $P(h_l)$ is the implicit probability model learned by the auto-encoder
which sits on top of $h_l$.

The basic reason why a denoising auto-encoder learns to map its input to a
nearby more probable configuration is that it is trained to do so: it is
trained with (input,target) pairs in which the input is a corrupted version
of the target, and the target is a training example. By definition,
training examples are supposed to have high probability in average (this is
what maximum likelihood is trying to achieve), and a random move around a
high probability configuration is very likely to be a low probability
configuration. In the maximum likelihood setup, we are trying to force the
model to put a high probability on the examples and a small probability
everywhere else. This is also what is going on here: we are telling the
auto-encoder that it should map points that are not training examples to a
nearby training example. This mental picture also highlights two things:
(1) the auto-encoder training criterion is more local than maximum
likelihood (if the noise level is small, it only sees configurations in the
neighborhood of the data), but higher noise levels should mitigate that,
and (2) if we increase the noise level we will make the model do a better
job at killing off spurious modes (configurations that are probable under
the model but should not), however the model then might start fuzzying its
reconstruction because the same corrupted configuration could then be
reached from many training examples, so the reconstruction would tend to be
somewhere in the middle, thus filling the convex between neighboring
training examples with high probability values. The latter observation
suggests the nearest-neighbor training procedure sketched in
Section~\ref{sec:nearest-neighbor} to avoid that problem. Another
(orthogonal) solution to spurious modes, that was previously
proposed~\citep{Bengio-et-al-NIPS2013-small} and works well, is the {\em
  walk-back procedure}: the injected noise is not completely random but
along paths following reconstruction, i.e., we let the learner go from a
training example towards a spurious mode by iterative encode/decode steps
and then punish it by telling it to reconstruct the starting training
example (this is similar to Contrastive Divergence).

Although we can consider $\hat{h}_l - h_l$ as a proxy for $\frac{\partial \log P(h_l)}{\partial h_l}$
and providing a vector field (a vector for each point $h_l$),
keep in mind that this vector is not guaranteed to be a proper gradient, in the sense that
integrating it through different paths could give slightly different results. Only in the asymptotic
non-parametric limit is $\hat{h}_l - h_l$ converging to a gradient field.
The previous analysis of denoising auto-encoders~\citep{Alain+Bengio-ICLR2013,Bengio-et-al-NIPS2013-small}
clearly shows that the auto-encoder implicitly estimates a distribution $P(H_l)$
so as to match it as well as possible to its training distribution
$Q(H_l)$. This is consistent with even earlier
results showing that the denoising criterion is a regularized
form of score matching~\citep{Vincent-2011} called
denoising score matching~\citep{Swersky-ICML2011}. It is also
consistent with a geometric intuition explained in 
\citet{Bengio-Courville-Vincent-TPAMI2013} suggesting that
if the auto-encoder is regularized (in the sense that it is prevented
from perfectly reconstructing every possible input), then the output
of the encoder will be locally most (or even only) sensitive to the
probable variations of the data around the input, and only reconstruct
well for inputs that are near these highly probable regions
(manifolds). Indeed, the auto-encoder only has limited representation
resources (this is due to the regularization, contraction of the
encoder and decoder functions) and in order to minimize reconstruction
error it must use this capacity where it is really needed, i.e.,
to capture the variations in regions of high density of the
data distribution, while ignoring variations not present in the data,
by mapping unlikely input configurations towards nearby more likely
configurations.

Based on this intuition, we conjecture that the above result regarding
the meaning of $\hat{h}_l-h_l$ can be generalized to other settings,
e.g., where $h_l$ is discrete or where a penalty other than the squared
error is used to push $\hat{h}_l$ towards $h_l$ when training the auto-encoder.
In support of this conjecture, there is the work on {\rm ratio matching}~\citep{Hyvarinen-2007},
which generalizes {\rm score matching} to models on binary vectors.
Ratio matching constructs an energy function (an unnormalized probability
model) model of the vector of input bits by relying only in its training
criterion on the relative energy associated with the observed $x$
and a ``noisy'' $x$ obtained by flipping one of the bits.
It turns out that the ratio matching criterion can also be rewritten
in a way similar to an auto-encoder reconstruction error (Eq. 3 of
~\citet{Dauphin+Bengio-NIPS2013}), with the $i$-th input being
masked when reconstructing it (similarly to pseudo-likelihood).
This is similar to the denoising auto-encoder but where the corruption
noise consists in hiding only one of the bits at a time.
This also suggests (although in a specialized setting) that the
reconstruction probability associated to each input bit is ``pointing''
in the direction of a more probable value, given the input, but
that remains to be proven in the general case.

\begin{figure}[ht]
\begin{center}
\ifarxiv
\centerline{\includegraphics[width=0.5\columnwidth]{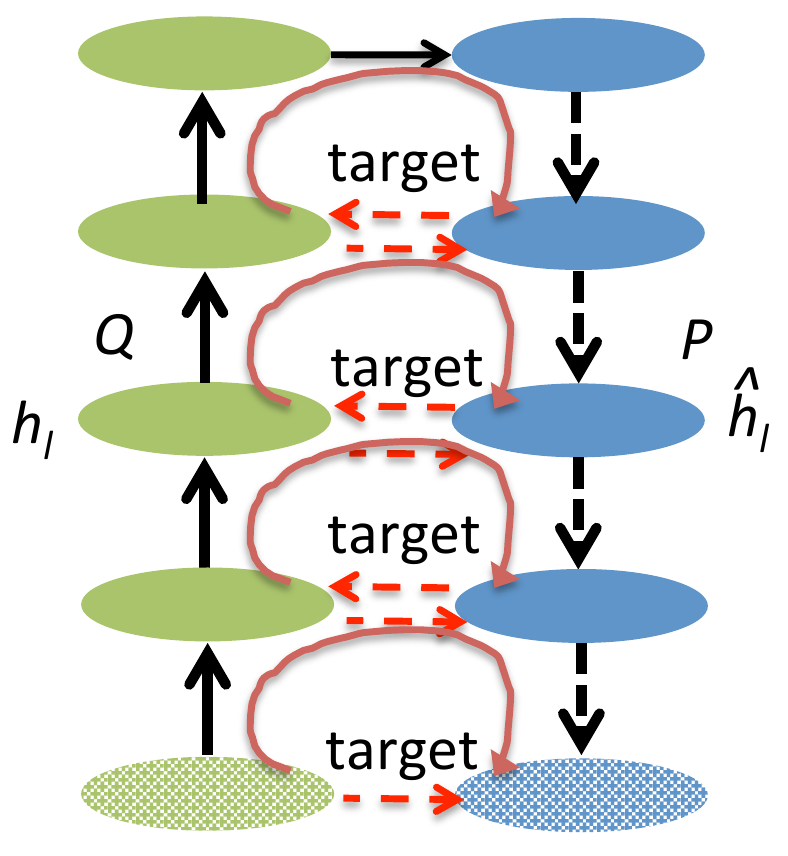}}
\else
\centerline{\includegraphics[width=0.5\columnwidth]{figures/short-and-long-loops.pdf}}
\fi
\caption{The training procedure involves running both a long-loop encode/decode path
(in thick black, bold going up, producing representations $h_l$,
and dashed going down, producing reconstructions $\hat{h}_l$) 
and all the short-loop encode/decode
paths associated with each layer (in red, curved). The representations $h_l$
and reconstructions $\hat{h}_l$ basically become targets for each other,
while each layer is trained to be a good auto-encoder.
}
\label{fig:generative-training}
\end{center}
\end{figure} 

\subsubsection{Putting It All Together: Architecture and Training Scheme}

The above allows us to propose an architecture and training scheme
for deep generative models that are also deep auto-encoders. In
Algorithm~\ref{alg:multi-KL} the general structure of the computation
and training loss is outlined, while a possible back-prop free implementation
using a variant of the recirculation algorithm
(for training a single-layer auto-encoder without back-prop)
is sketched in Algorithm~\ref{alg:generative-training} and
Figure~\ref{fig:generative-training}.

\begin{algorithm}[ht]
\caption{Proposed Architecture and Training Objective.
Each of the $L$ layers parametrizes an encoder $f_l$ and a decoder $g_l$.
The objective is to make
  (1) each layer a good layer-wise denoising auto-encoder, (2) every ``long loop''
  (going all up from $h_l$ and down to $\hat{h}_l$) also a good denoising
  auto-encoder, (3) every encoder produce output close the reconstruction of
  the noise-free long-loop auto-encoder above it, (4) every layer
  (but especially the top one) produce an appropriate amount of noise,
  just enough to make the decoder sufficiently contractive but not hurt
  reconstruction error too much. The top layer may be allowed to 
  have more noise and a different prior
  (e.g. as in~\citet{Kingma+Welling-ICLR2014}). $\alpha$=1 is
  reasonable choice, and controls the trade-off between reconstruction
  accuracy and the pressure to make $Q(h)$ simple.
}
\label{alg:multi-KL}
\begin{algorithmic}
\STATE Sample training example $h_0=x \sim Q(X)$
\STATE Cost $C \leftarrow 0$
\FOR {$l=1 \ldots L$}
  \STATE $h_l \sim Q(H_l | h_{l-1})$, with a deterministic component $f_l(h_{l-1})=E_Q[H_l | h_{l-1}]$ 
   and noise injected with entropy $\log \sigma_l$, e.g., $h_l = f_l(h_{l-1}) + \sigma_l \xi_l$ and $\xi_l$ is a Normal$(0,I)$ 
   random vector.
  \STATE $C \leftarrow C - \log \sigma_l - \log P(x | h_l)$ where the second term corresponds to minimizing the denoising
  reconstruction error through the $x \rightarrow h_l \rightarrow \hat{x}$ loop.
\ENDFOR
\STATE $\hat{h}_L = h_L$
\FOR {$l=L \ldots 1$}
  \STATE $\hat{h}_{l-1} = g_l(\hat{h}_l)$ 
  \STATE Ideally we want $C \leftarrow C - \log P(h_{l-1})$ but this is not generally tractable. However,
  the gradient of this term w.r.t. the upper parts of the model
  corresponds to minimizing denoising reconstruction error through the $h_{l-1} \rightarrow h_L \rightarrow \hat{h}_{l-1}$
  loop, and the gradient of this term w.r.t. lower parts of the model is obtained by using
  $\hat{h}_{l-1}-h_{l-1}$ as a proxy for $\frac{\partial \log P(h_{l-1})}{\partial h_{l-1}}$.
  Hence, instead of the above, for the intermediate layers, 
  we have $C \leftarrow C - \log P(h_{l-1} | h_L)$ for training
  the upper loop denoising auto-encoder (gradient into the conditional distribution over $H_{l-1}$
  given $h_L$, which predicts $\hat{h}_{l-1}$ as expected value) and also for
  training the lower level encoders (gradient into the ``observed'' $h_{l-1}$ which is produced
  by lower-level $f$'s and their parameters). For the top layer, the prior may be computable analytically,
  in which case one can directly maximize $\log P(h_{l-1})$ wrt both $Q$ and the prior itself.
\ENDFOR
\STATE Update parameters by performing a stochastic gradient step w.r.t. $C$.
\end{algorithmic}
\end{algorithm}

\begin{algorithm}[ht]
\caption{Sketch of tentative backprop-free implementation.
Each of the $L$ layers parametrizes an encoder $f_l$ and a decoder $g_l$.
The auto-encoders at each layer must be good enough for the top-level
reconstruction signals to provide good targets to lower levels.
When a target on some value (the output of an encoder $f_l$ or
decoder $g_l$) is specified, it may be used to provide a gradient
signal for that encoder or decoder. The top-level auto-encoder's
ability to be a good generative model could be improved in
various ways, e.g. using the walk-back procedure~\citep{Bengio-et-al-NIPS2013-small}
in which we let it go up and down several times with noise injected and
then drive the reconstruction $\hat{h}_{L-1}$ towards either $h_{L-1}$
or a nearby training example's $h_{L-1}$ representation. This is meant as 
an a back-prop free approximation of Algorithm~\ref{alg:multi-KL}.
}
\label{alg:generative-training}
\begin{algorithmic}
\STATE Sample training example $h_0=x \sim Q(X)$
\FOR {$l=1 \ldots L$}
  \STATE $h_l = f_l(h_{l-1})$
  \STATE $h_{l-1}$ is a target for $\tilde{h}_{l-1}=g_l(corrupt(h_l))$ (especially for updating $g_l$)
  \STATE $h_l$ is a target for $f_l(\tilde{h}_{l-1})$ (especially for updating $f_l$)
\ENDFOR
\STATE $\hat{h}_L = h_L$
\FOR {$l=L \ldots 1$}
  \STATE if $l<L$, $\hat{h}_l$ is a target for $h_l$ (especially for updating $f_l$)
  \STATE $\hat{h}_{l-1} = g_l(\hat{h}_l)$ 
  \STATE $h_{l-1}$ is a target for $g_l(corrupt(\hat{h}_l))$ (especially for updating $g_l$)
\ENDFOR
\end{algorithmic}
\end{algorithm}

The corruption helps to make both the
encoders and decoders contractive. A reasonable choice for the level of corruption
is given by the nearest-neighbor distance
between examples in the corresponding representation. In this way, the
``empty ball'' around each training example is contracted towards that example,
but we don't want to contract one training example to its neighbor.
The corruption
may actually not be necessary for the lower layers encoders because they are
regularized by the layers above them, but this corruption is certainly necessary
for the top layer auto-encoder, and probably for the lower-level decoders
as well.

Notice however how the up-going $h_l$ and down-going $\hat{h}_l$ paths (which we call
the {\em long loop}) are free of noise.
This is to obtain as clean as possible of a target. The mathematical
derivation of $\hat{h}_l - h_l$ as an estimator of $\frac{\partial \log P(h_l)}{\partial h_l}$
(up to a constant proportional to the corruption level) relies on the noise-free
reconstruction~\citep{Alain+Bengio-ICLR2013}.

In Algorithm~\ref{alg:generative-training}, ``A is a target of $F(b)$'' means that $F(b)$ should
receive a gradient pushing it to produce an output closer to $A$.
For example, with squared error, the gradient on $F(b)$ would be $F(b)-A$.
If backprop is not used between layers, an objective that guides the design
proposed here, then only the parameters of $F$ are updated.

Note that the way in which we propose to make a good denoising auto-encoder out of
any stack of auto-encoders starting above layer $h_l$, i.e., capturing $P(H_l)$,
is slightly different from the traditional denoising auto-encoder training
procedure. It is motivated by the need to train {\em all of them} (for all $l$)
at once, and by the objective to make {\em both the encoders and decoders
contractive}, whereas the traditional objective only makes their
composition (decode(encode(input))) contractive. The last motivation is that we want
the above training to have a chance to work {\em even without backprop}
across layers, i.e., assuming that a training signal on some output 
of an up-going encoder $f_l$ or down-going decoder $g_l$ 
is only used for training that layer-wise encoder or decoder.
The idea that we propose in order to train each layer's auto-encoder
without having to use backprop to propagate reconstruction error
into the encoder (through the decoder) follows from the
recirculation algorithm of ~\citet{Hinton+McClelland-NIPS1987}.
At each layer we consider an up-down-up (encode-decode-encode)
step where the decoder gets its reconstruction target from the 
initial input and the encoder gets its reconstruction target
from the initial output (of the encoder, before being decoded
and re-encoded).
All this is achieved by making each layer-wise auto-encoder 
pair ($f_l$,$g_l$) a good auto-encoder both ways, but only
for the data that matters, i.e., ideally coming from $Q$.

Note that there are two ways in which each layer-wise encoder becomes contractive:
(1) because of the pull towards modes of $P(H_l)$ in the line
``$\hat{h}_l$ is a target for $h_l$'', and
(2) because of the corruption noise in $\tilde{h}_{l-1}=g_l(corrupt(h_l))$ which is used in
``$h_l$ is a target for $f_l(\tilde{h}_{l-1})$''.
We may want to add more targets to make sure that $f_l$ is also
contractive around samples from $Q(H)$, e.g., 
``$h_l$ is a target for $f_l(corrupt(h_{l-1}))$''.
Similarly, it might be useful to make the layer-wise auto-encoders
good at auto-encoding not just the samples from $Q(H)$ but also
those from $P(H)$. This may be useful
in order to map samples in the neighborhood of $\hat{h}_l$ towards $\hat{h}_{l-1}$.
This would be good because an imperfect $P(H_l)$ (which does not perfectly
imitate $Q(H_l)$) will tend to be less peaky (have more entropy, be flatter)
then $P(H_l)$, i.e., it will tend to sample points in the neighborhood
of those that are likely under $Q(H_l)$, and we would like $g_l$ to map
these ``mistakes'' back towards the ``good values'' $h_l$.

\subsubsection{Backprop or No Backprop?}

In Algorithm~\ref{alg:generative-training}, we have a reconstruction
target $h_l$ for the upper auto-encoder and ``matching'' target
$\hat{h}_l$ for the encoder $\tilde{f}_l(x)$. In traditional auto-encoder
and neural network settings, such targets would be back-propagated
as much as possible to provide a signal for all the parameters that
influence the mismatch between $h_l$ and $\hat{h}_l$. We conjecture
here that it is not necessary to backprop all the way thanks to
the particular structure of the deep auto-encoder and the way
it is otherwise trained. We provide a justification for this
conjecture below.

\begin{figure}[ht]
\ifarxiv
\centerline{\includegraphics[width=0.5\linewidth]{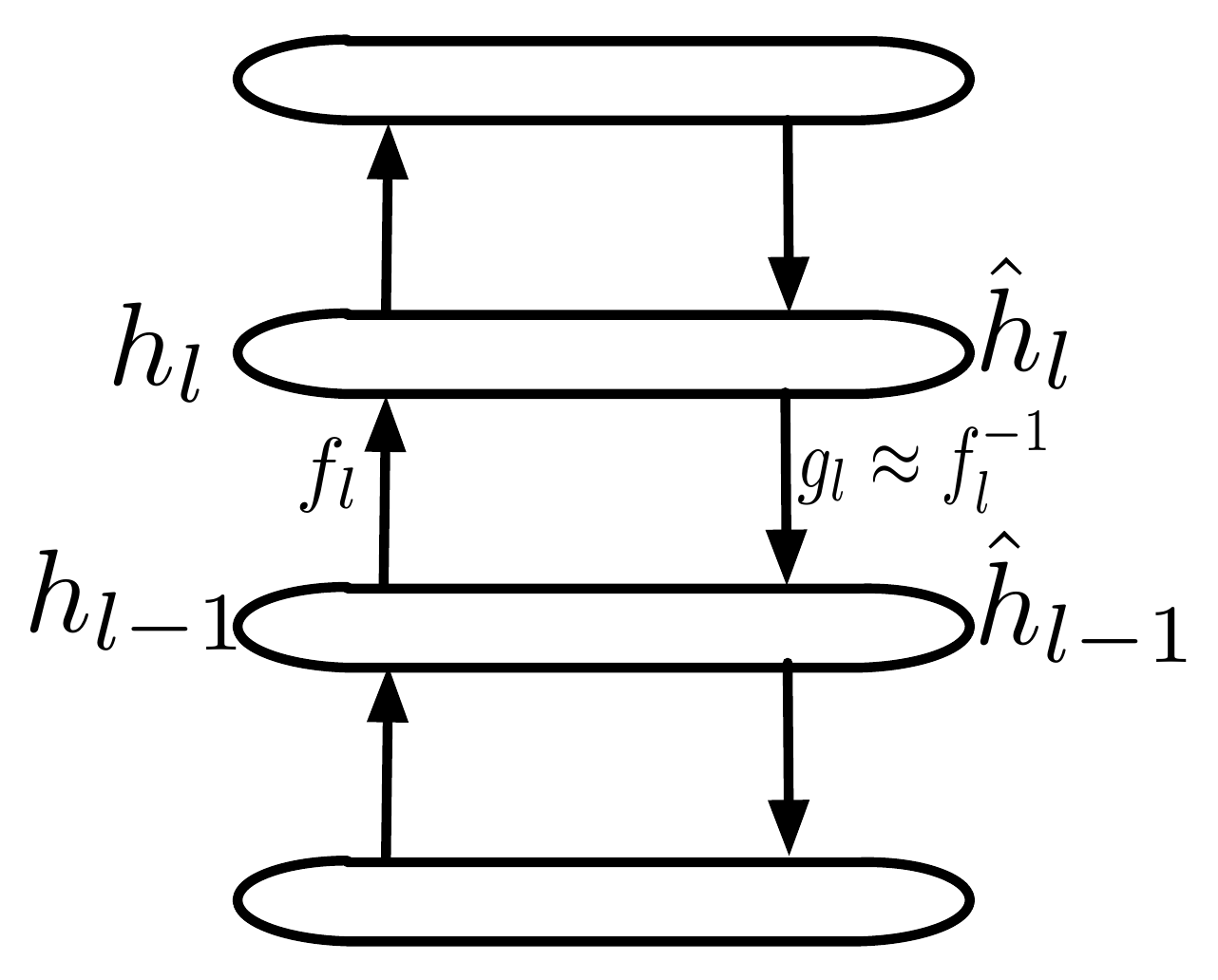}}
\else
\centerline{\includegraphics[width=0.5\linewidth]{figures/targetprop-layers.pdf}}
\fi
\caption{Illustration of auto-encoders allowing good targets at level $l$
to be propagated into good targets at level $l-1$ simply because decoder $g_l$ is
approximately inverting the encoder $f_l$. If $\hat{h}_l$ is a good target for $H_l$, around $h_l$,
i.e., if cost($H_l=\hat{h}_l) <\,$cost$(H_l=h_l)$ and $f_l(g_l(\hat{h}_l))\approx \hat{h}_l$,
i.e., the encoders are good inverses of each other in the neighborhood of the
reconstructions $\hat{h}_l$, then similarly we obtain that cost($H_{l-1}=\hat{h}_{l-1})<\,$cost$(H_{l-1}=h_{l-1})$,
i.e., the reconstructed $\hat{h}_{l-1}$ is also a good target for $H_{l-1}$.
}
\label{fig:targetprop-layers}
\end{figure}

The main justification for the usefulness of auto-encoders for
target-propagation arises out of the following observation.
If $\Delta h_l = \hat{h}_l - h_l$ is a vector indicating
a small change of $h_l$ that would increase some cost criterion
and with $h_l=f_l(h_{l-1})$ and $g_l$ roughly an inverse of
$f_l$ in the neighborhood of $h_l$, i.e., $f_l(g_l(\hat{h}_l))\approx \hat{h}_l$,
then a way to obtain $H_l=\hat{h}_l$ is to set $H_{l-1}=g_l(\hat{h}_l)$,
since then we get $H_l = f_l(g_l(\hat{h}_l)) \approx hat{h}_l$ as desired.
This is illustrated in Figure~\ref{fig:targetprop-layers}. More formally,
if $\hat{h}_l$ gives better outcomes then $h_l$, i.e.,
cost$(H_l=\hat{h}_l)<\,$cost$(H_l=h_l)$ and $\hat{h}_l$ is in
the neighborhood of $h_l$, i.e., $||h_l - \hat{h}_l|| < \epsilon$,
and $g_l(f_l(\hat{h}_l))=\hat{h}_l$, then $\hat{h}_{l-1}$ also gives
better outcomes than $h_{l-1}$, i.e.,
cost$(H_{l-1}=\hat{h}_{l-1})=\,$cost$(H_l=\hat{h}_l)<\,$cost$(H_l=h_l)=\,$cost$(H_{l-1}=h_{l-1})$
 and $\hat{h}_{l-1}$ is also in
the neighborhood of $h_{l-1}$, i.e., $||h_l - \hat{h}_l|| < L \epsilon$, where
$L$ is a bound on the derivatives of $f_l$ and $g_l$.
We can also obtain this by asking what would be a good target $\hat{h}_{l-1}$
for $h_{l-1}$, given a good target $\hat{h}_l$ for $h_l$:
it would one such that it gives the same cost as $\hat{h}_l$,
i.e., an $\hat{h}_{l-1}$ such that $f_l(\hat{h}_{l-1})=\hat{h}_l$.
If $f_l$ inverts $g_l$, then we can thus obtain that ideal target
via $\hat{h}_{l-1}=g_l(\hat{h}_l)$, since it means that
$f_l(\hat{h}_{l-1})=f_l(g_l(\hat{h}_l))\approx \hat{h}_l$.

Hence, if we set a small target change upon some top layer,
the reconstruction path computes corresponding small target changes
for all the layers below. Clearly, this generalizes the notion
of back-propagation in a way that may extend to non-differential
changes, since the above logic works for $\Delta h_l$ that is
not infinitesimal, and $h_l$ may even be discrete. The crucial
requirement is that $f_l$ be an inverse of $g_l$ in the neighborhood
of the data (the values $h_l$ obtained by feedforward computation
from data examples).

Another interesting observation is that we are imposing the $KL$ divergence criterion
{\em at every layer}. First, consider the training
signal on the encoder $\tilde{f}_l=f_l \circ \tilde{f}_{l-1}$.
The lower-level deep encoder $\tilde{f}_{l-1}$ is already receiving a training signal 
towards making it transform
$Q(X)$ into a distribution $Q(H_{l-1})$ that matches $P(H_{l-1})$ well,
while $P(H_l)$ is related to $P(H_{l-1})$ through the
decoder $g_l$, which maps samples from $P(H_l)$
into samples of $P(H_{l-1})$. Each layer of the encoder
is trying to transform its distribution into one that
is going to be easier to match by the upper layers,
flattening curved manifolds a bit better (keep in
mind that a completely flat manifold can be modeled
by a single linear auto-encoder, i.e., PCA). Hence
we conjecture that it is sufficient to only modify
$f_l$ (and not necessarily $\tilde{f}_{l-1}$) towards
the target $\hat{h}_l$. This is because $\tilde{f}_{l-1}$
itself is going to get its own target through the target
propagation of $\hat{h}_l$ into $\hat{h}_{l-1}$. This is
analogous to what happens with backprop: we use the
gradient on the activations of an affine layer $l$
to  update the weights of that layer, and it is the
back-propagation of the gradient on layer $l$ into
the gradient on the layer $l-1$ that takes care of
the lower layers. In a sense, back-propagating these
targets through more than one layer would be redundant.

Second, consider the training signal on the upper auto-encoder. The
consistency estimation theorem for denoising auto-encoders presented in
\citet{Bengio-et-al-NIPS2013-small} only requires that the last step of the
decoder be trained, so long as it has enough capacity to map its input to
its target. Furthermore, if the auto-encoder layer taking $h_{l+1}$ in
input is presumably already doing a good job (both in the sense
of minimizing reconstruction error and in the sense that $P(H_{l+1})$
and $Q(H_{l+1})$ are close to each other), then $g_{l+1}$ only
needs to learn to prefer the actually sampled ``data'' $h_l$ to other
nearby values, i.e., contracting towards the values it sees as
training targets.

It might still be the case that back-propagating all the way
further helps Algorithms~\ref{alg:multi-KL} and ~\ref{alg:generative-training} to converge
faster to a good model of the raw data $x$, but the main conjecture
we are making is that by providing a training signal at each
layer, the proposed training scheme will be less prone to the
training difficulties encountered when training very deep non-linear networks
with back-propagated gradients. Where we expect the back-propagation
through layers to give more of an advantage is in the supervised
scenario of Algorithm~\ref{alg:supervised-training}, below.

What we hope is that target propagation can side-skip the difficulties that
arise when using back-propagation for credit assignment, when the
dependencies to be captured are highly non-linear. We already know that
strong non-linearities arise out of the composition of many layers in a
deep net or many steps in a recurrent net and make it difficult to learn by
back-propagated gradients because the gradients tend to be either very
small (gradient vanishing problem) or very large (gradient explosion
problem). This has been well studied in the case of recurrent
networks~\citep{Hochreiter91-small,Bengio-trnn93-small,Hochreiter+Schmidhuber-1997},
and more recently by~\citet{Pascanu+al-ICML2013-small}.  However, more
generally, what to do when the non-linearities are so strong that the
derivatives are nearly or actually 0 or infinite? This is what happens with discrete
activation functions.

A major advantage of not completely relying on backprop to perform
credit-assignment through strong non-linearities is therefore that we can now
consider the case where the hidden layer representations are discrete
or combine discrete and continuous units at each layer. An
advantage of discrete representations is that they are
naturally {\em contractive}, since many input values can potentially
be mapped to the same output value due to the discretization step.
For the same reason, a discrete auto-encoder should be 
naturally ``error-correcting''. Discrete (but distributed) representations
might also be the most natural way to represent some aspects of the
data that are discrete (like categories), or input data that
are themselves discrete, such as language data. On the other hand, some
types of data and underlying factors are more naturally represented
with real values, so we should probably design systems that can capture
both continuous and discrete types.

\subsubsection{Sampling From the Model}

The training criteria (one for each layer) that are being optimized
suggest that we can sample from the model in many ways, all providing
a possibly different estimator of the data
generating distribution (if the encoder/decoder pairs are powerful
enough)\footnote{We have not proven that here, but we conjecture
that a consistency theorem can be proven, and its proof would hinge
on having enough layers to make sure that $P(H_l)$ approaches $Q(H_l)$,
for all layers. There are already proofs that deep and thin generative networks
with enough layers can approximate any distribution~\citep{Sutskever+Hinton-2008}. 
With enough layers, we conjecture that one should be able to map any
distribution to a factorial one.}.

Hence, in principle (due to the fact that we are minimizing $KL(Q(X,H_l)||P(X,H_l))$,
Eq.~\ref{eq:Ql}, or a bound on the log-likelihood),
 we can choose any level $l$ and generate a sample of $x$ as follows:
\begin{enumerate}
\item Sample $h_l$ from $P(H_l)$.
\item Sample $x$ from $P(X|h_l)$, i.e., $x=\tilde{g}_l(h_l)$.
\end{enumerate}
The first step involves sampling from the deep denoising auto-encoder
that sees $h_l \sim Q(H_l)$ as training data (and goes up to $h_L$ to
encode $h_l$). As demonstrated
in~\citet{Bengio-et-al-NIPS2013-small}, that can be achieved
by running a Markov chain where at each step we corrupt the previous
MCMC sample, encode it, and decode it. In general one should also
add noise proportional to the entropy of $P(H_l | h_L)$.
We have assumed that decoders were ``almost perfect'', i.e., that
the entropy of $P(H_l | h_L)$ is zero, in which case no noise
would need to be added. In practice, during training, there will be some
residual reconstruction error even when $h_l \sim Q(H_l)$. In that
case, it might be advantageous to sample from $P(H_l | h_L)$ in
the reconstruction step, i.e., add the appropriate noise. Note how the
effect of that noise is of smearing the distribution one would get
otherwise (convolving it with the noise distribution) so as to make
sure to include in the support of $P(H_l)$ the examples sampled from $Q(H_l)$.

However, as noted in~\citet{Alain+Bengio-ICLR2013}, one potential issue
with sampling from denoising auto-encoders is that if the amount of
corruption is small, then the chain mixes very slowly, and if it is large,
then the reconstruction distribution might not be well approximated by a
unimodal reconstruction distribution (which is what we are advocating here,
since we assume that the decoder is almost deterministic). What may save the day,
as argued in~\citet{Bengio-et-al-ICML2013} and
\citet{Bengio-et-al-NIPS2013-small}, is that mixing tends to be much easier
when done at higher levels of representation. In the case of this
paper, this is readily done by sampling at the top level, i.e., 
the generative procedure is summarized in Algorithm~\ref{alg:generate}.
In fact, if the top-level auto-encoder is {\em linear}, then its prior
is a Gaussian one (see Section~\ref{sec:linear-top}), and we can sample analytically, not requiring
a Markov chain. Similarly, if the top-level auto-encoder is an element-wise
auto-encoder (i.e., each dimension is auto-encoded separately), this really
corresponds to a factorial distribution, and again we can sample analytically
without requiring a Markov chain (see Section~\ref{sec:factorial-top}).
Another interesting direction of investigation is to replace the
reconstruction criterion of the penultimate level $h_{L-1}$ by one
in which one only tries to reconstruct to a training set near neighbor.
This idea is expanded in Section~\ref{sec:nearest-neighbor} below.

\begin{algorithm}[ht]
\caption{Generative Procedure Associated with the Training Scheme in Algorithm~\ref{alg:multi-KL}.
The corruption injected at the top controls the amount of mixing, but the default
level should correspond to sampling from $Q(H_L|h_{L-1})$.}
\label{alg:generate}
\begin{algorithmic}
\STATE \hspace*{3cm}{\sl First, sample $h_{L-1}$:}
\STATE Initialize $h_{L-1}$ randomly (or from a recorded $h_{L-1}$ which was sampled from $Q(H_{L-1})$).
\FOR {$k=1 \ldots K$}
  \STATE $h_{L-1}=g_L(corrupt(f_L(\hat{h}_{L-1})$
\ENDFOR
\STATE \hspace*{3cm}{\sl Second, map $h_{L-1}$ to $x$:}
\FOR {$l=L-1 \ldots 1$}
  \STATE $h_{l-1} = g_l(h_l)$
\ENDFOR
\STATE {\bf Return} $x=h_0$
\end{algorithmic}
\end{algorithm}

\subsubsection{Allowing the Top-Level to Mix Well with Nearest-Neighbor Reconstruction}
\label{sec:nearest-neighbor}

In order to allow the top-level to mix well while allowing the reconstruction distribution
to be unimodal and factorial (which is what a deterministic reconstruction really is),
we propose to consider training the top-level denoising auto-encoder or GSN with a criterion
that is different from the usual reconstruction criterion. This follows from an idea
jointly developed with Laurent Dinh~\footnote{Personal communication}. 
We call this new criterion the nearest-neighbor reconstruction criterion.

The motivation is that if we inject a lot of noise in the auto-encoder, it
will be impossible for the decoder to perfectly reconstruct the input, and
that will force it to have a high entropy (sampling from $P(X|h)$ will add
a lot of noise). However, in order for each stochastic
encode/decode step to correspond to a transition of a Markov chain that
estimates the data generating distribution as its stationary distribution,
it is enough that the decoder deterministically maps the top-level code 
towards the nearest ``mode'' of the data generating distribution, i.e.,
towards the nearest training example. This can be formalized as follows,
denoting $A$ for the stochastic transition operator implemented by a
stochastic denoising auto-encoder and $Q(X)$ the data generating distribution
(in our case we will apply this to the top level of representation).
A sufficient condition for $A$ to generate $Q(X)$ as its stationary distribution
is that $A$ mixes while applying $A$ to $Q(X)$ leaves $Q(X)$ unchanged.

Formally this means that the following criterion could be minimized:
\begin{equation}
\label{eq:KL-nn}
 KL(Q(X)|| A Q(X)) = -H(Q(X)) - E_{x \sim Q(X)} \log E_{x' \sim Q(X)} A(x|x')
\end{equation}
where $A Q(X)$ denotes the application of the linear operator $A$ to
the distribution $Q(X)$, and $A(x|x')$ is the probability that $A$ puts
on generating state $x$ when starting from state $x'$.

Intuitively, it means that we want any training example $x$ to be
reconstructible from at least some other example $x'$ on which some probable 
corruption (small amount of noise)
would have been applied (in our case, at the top level of representation).
The above criterion involves a double sum over the training data, which
seems to make it impractical. However, the inner expectation
may be approximated by its largest term, which is given by
the nearest neighbor of $x$ in representation space.
To see this clearly, first, let us introduce a noise source, which most conveniently
would be injected at the top level of the hierarchy, in some
latent variable $z$, i.e., $A$ is decomposed into the following
three steps: encode $x'$ into $z'=f(x')$, sample noise $\xi$ and add it to $z'$ with
$z=z'+\xi$, decode $z$ with $x=g(z)$. Furthermore, assume
the noise has a rapidly decaying probability as a function of $||z||$,
like the Gaussian noise, favoring the near neighbors of $x$ as
candidates for ``explaining'' it.

Let us consider an application of Jensen's inequality to Eq.~\ref{eq:KL-nn},
yielding the following upper bound training criterion:
\begin{equation}
 KL(Q(X)|| A Q(X)) \geq -H(Q(X)) - E_{x \sim Q(X)} E_\xi \log E_{x' \sim Q(X)} P(x|f(x')+\xi).
\end{equation}
We are considering the empirical distribution and assuming that $P(x|h)$ is nearly deterministic
and the noise $\xi$ has a rapidly decreasing probability in terms of $||\xi||^2$, such as the Gaussian noise.
Therefore the dominant term of the expectation over $x'$ will be max over $x'$, i.e., 
the nearest neighbor once we have added noise, and we get for each $x$  to minimize
\begin{equation}
  -  E_{\rm noise} \min_{x' \sim Q(X)} \log P(x|f(x')+{\rm noise}).
\end{equation}
which amounts to the following computations, for every example $x$:
\begin{enumerate}
\item Compute $f(x)$.
\item Sample and add noise $\xi$.
\item Look for nearest neighbor $f(x')$ of $f(x)-\xi$ among the training examples $x'$
(which can probably be approximated by looking only at the list of nearest neighbors of $x$ in input space).
\item Minimize reconstruction error of $x$ by the decoder $g(f(x')+\xi)$, both by 
changing the decoder $g$ and the encoder $f$. In the language of the previously
proposed algorithms, {\em the former means that $x$ is a target for $g(f(x')+\xi)$
and that $f(x)$ and $f(x')$ are targets for each other.}
\end{enumerate}

\section{Supervised or Semi-Supervised Learning Target Propagation}
\label{sec:supervised}

In this section, we explore how the ideas presented in the previous section
can be extended to provide a layer-local training signal in deep supervised
networks. This naturally provides a way to train a deep network in both
supervised and semi-supervised modes.  Each observed example is now assumed
to be either an $(x,y)$ input/target pair or a lone input $x$.

\begin{figure}[ht]
\begin{center}
\ifarxiv
\centerline{\includegraphics[width=0.5\columnwidth]{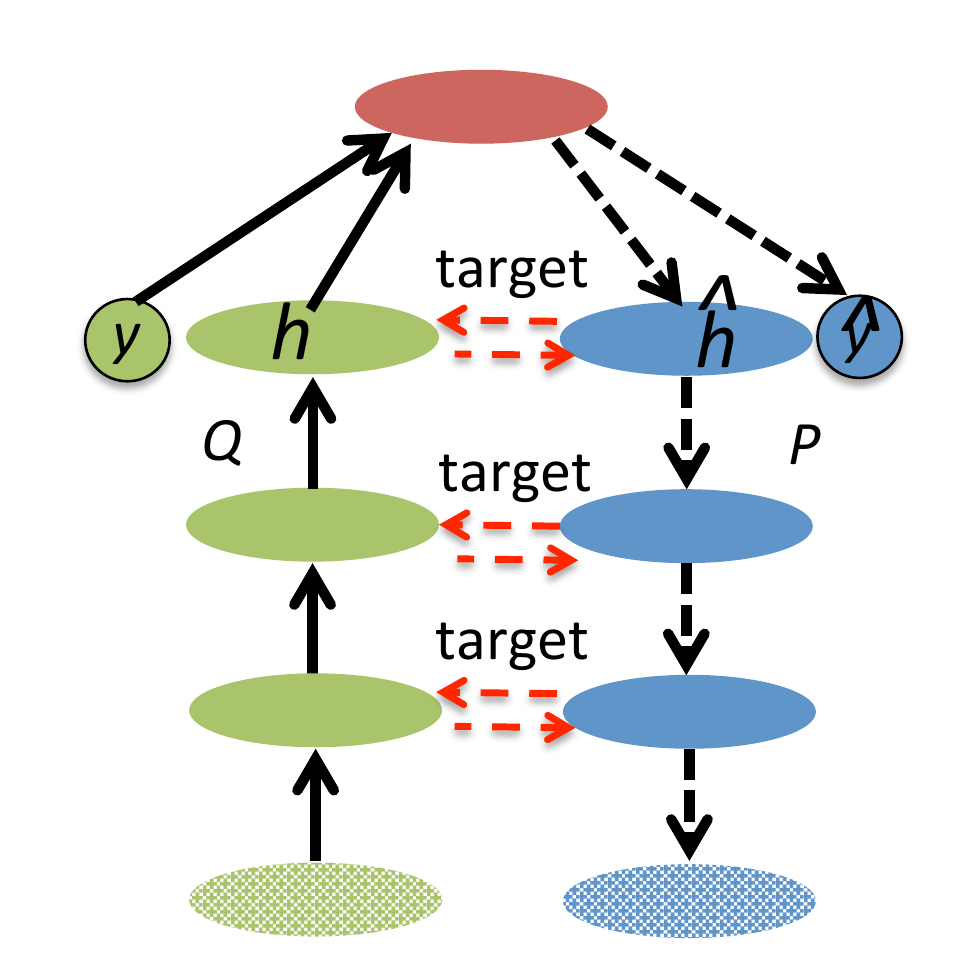}}
\else
\centerline{\includegraphics[width=0.5\columnwidth]{figures/supervised-training.pdf}}
\fi
\caption{Generalization to supervised training, with $y$ appearing as an extra
input of the top-level auto-encoder. The reconstruction $\hat{h}$ of $h$ when
$y$ is given provides a supervised training signal for changing the encoder $Q(h|x)$.
It can be made discriminant by considering as well the reconstruction obtained
when $y$ is missing.
}
\label{fig:supervised-training}
\end{center}
\end{figure} 

\subsection{Factorial Output Distribution}

We will first consider the case where $y$ is a low-dimensional object, such
as a category (for classification tasks) or a reasonably-sized real-valued
vector (for regression tasks) for which $P(Y|x)$ can be well approximated
by a factorial distribution, given $x$. In this case we can follow a strategy
initiated with Deep Belief Networks~\citep{Hinton06} and let the top-level
auto-encoder (instead of a top-level RBM) model the joint distribution of
$y$ and of $h_{L-1}$, the latter being the learned representation of $x$.

In this context, we consider a prediction of $y$ given $x$ to be simply
given by the reconstruction of $y$ given $h_L$. A denoising auto-encoder
(or more generally a Generative Stochastic
Network~\citep{bengio-et-al-ICML2014-small}) is naturally trained to
reconstruct some of its inputs given the others.
The ``missing'' inputs must be represented in some standard way that
allows the auto-encoder to distinguish the ``missing'' value from
likely values. When an input variable is discrete and encoded by a
one-hot vector (all zeros except a 1 at the $i$-th location),
a missing value can simply be represented by the vector of all zeros.
When an input variable is continuous, missingness can be represented
explicitly by a binary vector indicating which input is missing,
as in~\citet{Benigno-et-al-icml-2014}. The procedure for predicting $y$
given $x$ is presented in Algorithm~\ref{alg:supervised-prediction},
and illustrated in Figure~\ref{fig:supervised-training}.

The top-level encoder $f_L$ thus generally takes three arguments,
the input representation at the penultimate level, $h_{L-1}=\tilde{f}_{L-1}(x)$, 
the label $y$ (or 0 if $y$ is missing), and the mask $m$ (a bit indicating
whether $y$ is observed, $m=1$ or missing, $m=0$). The decoder $g_L$ takes $h_L$ in input
and predicts $h_{L-1}$ and $y$. We denote $g_L^y$ for the
part of $g_L$ that predicts $y$ and $g_L^h$ for the part of $g_L$ that predicts $h_{L-1}$.
When $y$ is a category (e.g., represented by a one-hot vector), then
as usual with neural networks, one would represent $g_L^y$ with a softmax layer
and use cross-entropy (i.e. negative log-likelihood of $y$ given $h_L$) as a ``reconstruction
error''.

\begin{algorithm}[ht]
\caption{Prediction Procedure Associated with the 
Supervised Target Propagation Training Scheme in Algorithm~\ref{alg:supervised-training}.
It takes $x$ as input and returns a prediction $\hat{y}$.}
\label{alg:supervised-prediction}
\begin{algorithmic}
\STATE $h_0=x$
\FOR {$l=1 \ldots L-1$}
  \STATE $h_l = f_l(h_{l-1})$
\ENDFOR
\STATE $h_L=f_L(h_{L-1},0,0)$
\STATE {\bf Return} $\hat{y}=g_L^y(h_L)$.
\end{algorithmic}
\end{algorithm}

This architecture, combined with the principle already presented in Algorithms~\ref{alg:multi-KL} or~\ref{alg:generative-training},
gives rise to the training procedure summarized in Algorithm~\ref{alg:supervised-training} for
the supervised or semi-supervised setups.

\begin{algorithm}[ht]
\caption{Target Propagation Training Procedure for Stacked Auto-Encoders in
  Supervised or Semi-Supervised Setups.  Once trained, such a deep network
  can be used for predictions following
  Algorithm~\ref{alg:supervised-prediction}. Like before, the objective is to make
  (1) each layer a good layer-wise denoising auto-encoder, (2) every ``long loop''
  (going all the way up from $h_l$ to $h_L$ and back down to $\hat{h}_l$) also a good denoising
  auto-encoder, (3) every encoder produce output close the reconstruction of
  the noise-free long-loop auto-encoder above it. In addition, the top level
  auto-encoder takes $y$ as an extra input. The gradient on $h_{L-1}$ can
  be made more discriminant as per Eq.~\ref{eq:hybrid-target}, with $\alpha=1$
  corresponding to the fully discriminant case and $\alpha=0$ for modeling
  the joint of $x$ and $y$.}
\label{alg:supervised-training}
\begin{algorithmic}
\STATE Sample training example, either $(x,y)$ (labeled) or $x$ (unlabeled).
\STATE $h_0=x$
\STATE $m=1$ if labeled, 0 otherwise (in which case $y$ can take an arbitrary value).
\FOR {$l=1 \ldots L-1$}
  \STATE $h_l = f_l(h_{l-1})$
  \STATE $h_{l-1}$ is a target for $\tilde{h}_{l-1}=g_l(corrupt(h_l))$ (especially for updating $g_l$)
  \STATE $h_l$ is a target for $f_l(\tilde{h}_{l-1})$ (especially for updating $f_l$)
\ENDFOR
\STATE $h_L = f_L(h_{L-1},m \times y, m)$
\STATE $(\hat{y},\hat{h}_{L-1}) = g_L(\hat{h}_L) = (g_L^y(\hat{h}_L),g_L^h(\hat{h}_L))$
\IF {labeled} 
  \IF {$\alpha>0$}
    \STATE Unsupervised reconstruction is $\tilde{h}_{L-1} = g_L^h(f_L(h_{L-1},0,0))$
    \STATE replace non-discriminant target $\hat{h}_{L-1}$ by hybrid target: \\
           $\hat{h}_{L-1} \leftarrow \hat{h}_{L-1} + (1-\alpha) (h_{L-1} - \tilde{h}_{L-1})$
  \ENDIF
  \STATE $y$ is a target for $g_L^y(corrupt(h_L)))$ (especially for updating $g_L^y$)
\ENDIF
\STATE $h_{L-1}$ is a target for $g_L^h(corrupt(\hat{h}_L))$ (especially for updating $g_L^h$)
\FOR {$l=L-1 \ldots 1$}
  \STATE $\hat{h}_l$ is a target for $h_l$ (especially for updating $f_l$)
  \STATE $\hat{h}_{l-1} = g_l(\hat{h}_l)$
  \STATE $h_{l-1}$ is a target for $g_l(corrupt(\hat{h}_l))$ (especially for updating $g_l$)
\ENDFOR
\end{algorithmic}
\end{algorithm}

A way to make sense of what is going on when training the encoders with the label $y$
being given is to consider what distribution is used to provide a target (i.e. a reconstruction) 
for $h_{L-1}$ in Algorithm~\ref{alg:supervised-training}. For this, we need to
generalize a bit the results in \citet{Alain+Bengio-ICLR2013} regarding the
estimation of $\frac{\partial \log P(h)}{\partial h}$ via the difference
between reconstruction $\hat{h}$ and input $h$.

Consider an auto-encoder modeling the joint distribution $P(C)=P(A,B)$ of variables $A$
and $B$, with $C=(A,B)$. For any given fixed $B=b$, the auto-encoder
therefore models the conditional distribution $P(A|B)$, as discussed
in~\citet{Bengio-et-al-NIPS2013-small}. As argued in more details
in Section~\ref{sec:discriminant}, the difference between the
reconstruction $\hat{a}$ of $a$ and $a$ itself is thus the model's
view of $\frac{\partial \log P(A=a|B=b)}{\partial a}$.

In our case, when we ``clamp'' $y$ to its observed value, what we get in
$\hat{h}_{L-1}-h_{L-1}$ is the top auto-encoder's estimated $\frac{\partial \log
  P(h_{L-1}|y)}{\partial h_{L-1}}$, when $h$ is continuous. In the discrete
case, we have argued above (but it remains to demonstrate formally) that 
$\hat{h}_{L-1}$ is an estimate by the model of a nearby most likely
neighbor of $h_{L-1}$.

Extending the argument to lower layers, we see that each $\hat{h}_l$
is an estimate by the upper layers of a value of the $l$-th level
representation that is near $h_l$ and that is more likely than
$h_l$, {\em given} $y$. This is the sense in which this procedure
is related to back-propagation and deserves the name of {\em target propagation}.

\begin{figure}[ht]
\begin{center}
\ifarxiv
\centerline{\includegraphics[width=0.5\columnwidth]{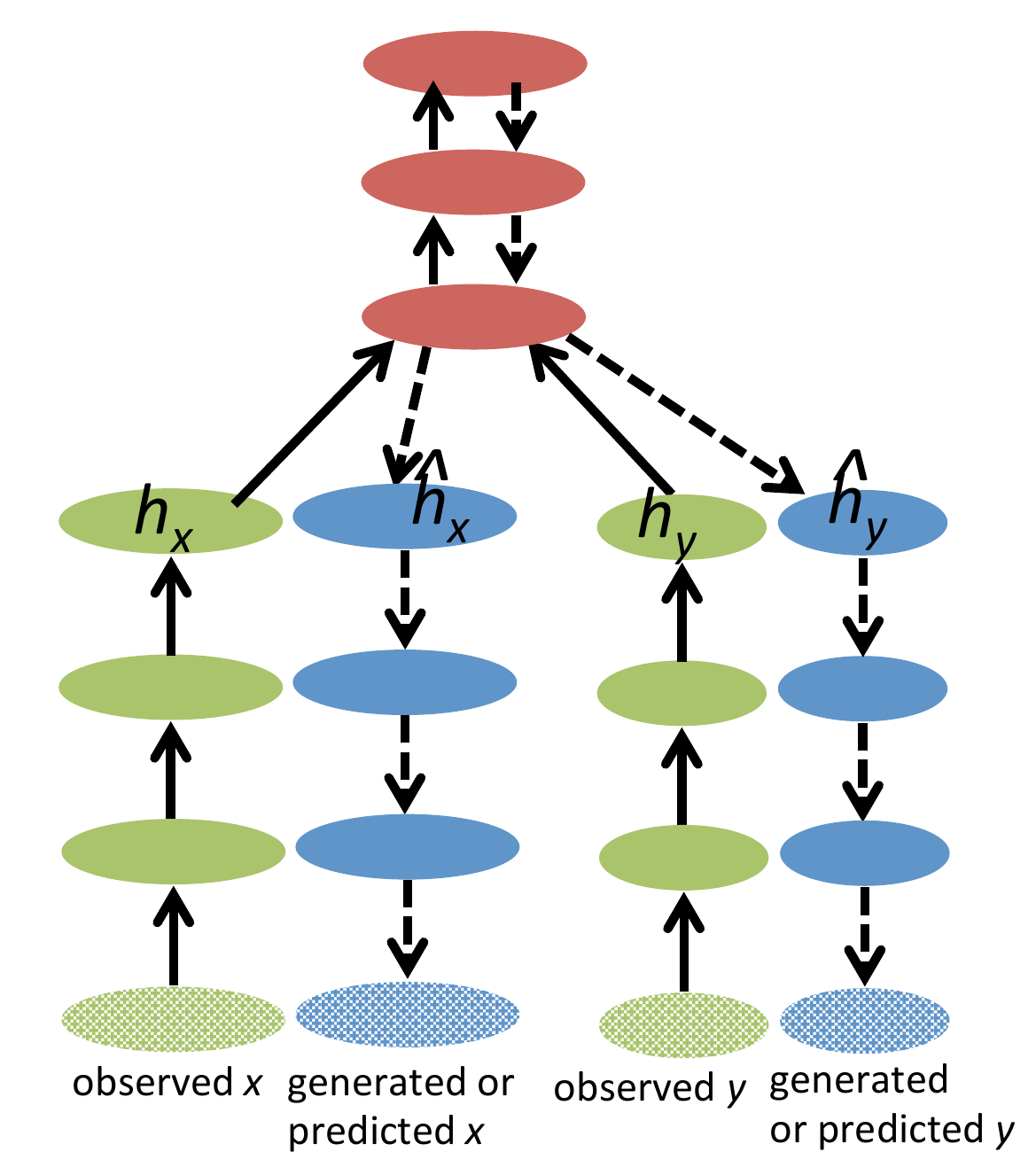}}
\else
\centerline{\includegraphics[width=0.5\columnwidth]{figures/multi-modal.pdf}}
\fi
\caption{Architecture associated with the structured output or
multi-modal cases (two modalities shown, here, $x$ and $y$). Each
modality has its own representation, but a top-level deep auto-encoder
learns to represent their joint or conditional distribution.
The top-level auto-encoder can be used to sample (MCMC) 
or predict (MAP) some modalities given others.
}
\label{fig:multi-modal}
\end{center}
\end{figure} 

\subsection{Structured Outputs}

If $y$ has a complicated non-factorial conditional distribution, given $x$,
then a simple deterministic function $g_L^y(h_L)$ to predict $E[Y|h_L]$ is not going to
be enough, and we need to find a more powerful way to capture $P(Y|x)$.

In that case, we can have two stacks of auto-encoders, one that mostly models
the $x$-distribution, $P(X)$, and one that mostly models the $y$-distribution,
$P(Y)$, but with the top-level codes \mbox{$h^x_{L-1}=\tilde{f}^x_{L-1}(x)$}
and $h^y_{L-1}=\tilde{f}^y_{L-1}(y)$ being such that their joint distribution
can easily be modeled by a top-level auto-encoder. Note that because this
joint distribution might be complex (and not fully captured by the transformations
leading from $x$ to $h^x_{L-1}$ and from $y$ to $h^y_{L-1}$), the top-level
auto-encoder itself will probably need to be a deep one (with its own
intermediate layers), rather than a shallow one (which should be appropriate
for straightforward classification problems). In other words, whereas
each stack computes useful features for $x$ and $y$ separately, useful
features for their joint distribution generally requires combining information from
both $x$ and $y$. When $y$ is an explicit one-hot, this does not seem necessary,
but in many other cases, it probably is. This architecture is illustrated
in Figure~\ref{fig:multi-modal}.

Sampling from $P(Y|x)$ proceeds as one would expect in a conditional
auto-encoder, i.e., compute $h^x_{L-1}$, consider it fixed (clamped),
and run a Markov chain on the top-level auto-encoder, resampling
only $h^y_{L-1}$ at each step. Then map $h^y_{L-1}$ to $y$ through
the deterministic $\tilde{g}^y_{L-1}$.

\subsection{Discriminant Training}
\label{sec:discriminant}

Let us consider more closely the training signal that is propagated
by the reconstruction targets in the case of supervised learning.

If we use the reconstruction $\hat{h}^x_{L-1}$ of $h^x_{L-1}$ by the top-level auto-encoder (possibly
a deep one in the structured output case) as target for $h^x_{L-1}$, then what
is really happening is that we are modeling the joint distribution
of $x$ and $y$. This can be seen by observing that in the continuous
case, the target minus the input is
\[
  \hat{h}^x_{L-1} - h^x_{L-1} \propto \frac{\partial \log P(h^x_{L-1}, h^y_{L-1})}{\partial h^x_{L-1}}.
\]
In the non-structured output case, this is simply
\[
  \hat{h}^x_{L-1} - h^x_{L-1} \propto \frac{\partial \log P(h^x_{L-1}, y)}{\partial h^x_{L-1}}.
\]

Because the normalization constant over $h^x_{L-1}$ does not depend of $h^x_{L-1}$,
we can equivalently write
\begin{align}
  \hat{h}^x_{L-1} - h^x_{L-1} \propto
   & \frac{\partial \log P(h^x_{L-1}, h^y_{L-1})}{\partial h^x_{L-1}} \nonumber \\
   =& \frac{\partial \log P(h^x_{L-1} | h^y_{L-1}) + \log P(h^y_{L-1})}{\partial h^x_{L-1}} \nonumber \\
   =& \frac{\partial \log P(h^x_{L-1} | h^y_{L-1})}{\partial h^x_{L-1}} \nonumber \\
   =& \frac{\partial \log P(h^x_{L-1} | y)}{\partial h^x_{L-1}}.
\end{align}
Therefore, we can view the training signal as maximizing the likelihood under the prior of $P$
for $h^x_{L-1}$, but conditioned by side information $y$.  This remark becomes useful
for the next two sections, when we consider other modalities, or the past or future
observations, as side information.

But if we want to perform {\em discriminant training} of a deep supervised network,
the conditional whose gradient we care about is $\frac{\partial \log P(y | h^x_{L-1})}{\partial h^x_{L-1}}$.
Fortunately, it can be easily computed:
\begin{align}
  \frac{\partial \log P(y | h^x_{L-1})}{\partial h^x_{L-1}} = &
\frac{\partial \log P(h^x_{L-1}, y) - \log P(h^x_{L-1})}{\partial h^x_{L-1}} \nonumber \\ 
 =&   \frac{\partial \log P(h^x_{L-1}, y)}{\partial h^x_{L-1}} - \frac{\partial \log P(h^x_{L-1})}{\partial h^x_{L-1}} \nonumber \\
\propto (\hat{h}^x_{L-1} - h^x_{L-1}) - (\tilde{h}^x_{L-1} - h^x_{L-1}) \nonumber \\
= \hat{h}^x_{L-1} - \tilde{h}^x_{L-1} \nonumber \\
\end{align}
where we have denoted $\tilde{h}^x_{L-1}$ the reconstruction of $h^x_{L-1}$ when $y$ is not provided
(which is obtained by setting the non-missingness mask of $y$ to 0, i.e., the auto-encoder
is modeling $x$ alone).
This gives the following {\em target} for $h^x_{L-1}$:
\begin{equation}
  {\rm discriminant \; target} = \hat{h}^x_{L-1} - \tilde{h}^x_{L-1} + h^x_{L-1}
\end{equation}
instead of $\hat{h}^x_{L-1}$. What it means is that we want to remove
from the reconstruction target (with $y$ given) the target that would have
been given if $y$ had not been given. 

Target propagation can then proceed as usual for the lower layers.

As in~\citet{Larochelle+Bengio-2008-small}, one can balance (in a
data-selected way) the objective of modeling the joint between $x$ and $y$
with the discriminant objective of modeling the conditional of $y$ given
$x$, by weighing the gradients associated with these different targets
appropriately. For example, with weight $\alpha \in [0,1]$ on the joint likelihood
and weight $(1-\alpha)$ on the discriminant likelihood, we obtain
the hybrid target
\begin{equation}
\label{eq:hybrid-target}
 {\rm hybrid \; target} = \hat{h}^x_{L-1} + (1-\alpha) (h^x_{L-1} - \tilde{h}^x_{L-1}).
\end{equation}

\section{Multi-Modal Modeling}

The supervised and semi-supervised setup described in the previous
two sections can easily be generalized to multi-modal modeling.
In particular, if there are two general modalities, then the
setup of the previous section for learning to represent $P(X,Y)$
and sampling $P(Y|x)$ can be trivially generalized to obtain
a way to sample $P(X|y)$. This is illustrated
in Figure~\ref{fig:multi-modal}.

If there are $N$ modalities $X^{(1)},\ldots X^{(N)}$, then the architecture can be naturally
generalized as follows. Have one stack of auto-encoders for
each modality, used to transform each modalities data $X^{(t)}$ into
a representation where the marginal $P(X^{(t)})$ can be captured easily
through some $P(H^{(t)})$, with $h^{(t)} = \tilde{f}^{(t)}_{L-1}(x^{(t)})$.
Then let $h_{L-1}=(h^{(1)}_{L-1}, \ldots h^{(2)}_{L-1}, h^{(1)}_{L-1})$
and model $h_{L-1}$ with another deep stack of generative auto-encoders.

During training, when a subset of modalities is available, encode
the missingness of any modality through a missingness mask $m^{(t)}$
for modality $t$, and use it to turn off the 
output of $\tilde{f}^{(t)}_{L-1}(x^{(t)})$ when $x^{(t)}$ is missing
(just as we did above when $y$ is missing, in the regular supervised case).
Train the top-level deep auto-encoder on the given examples,
possibly randomly hiding some subset of the observed modalities
so as to provide a target for $h^{(t)}$ for those modalities
that were available in the training example but that were hidden
at the input of the top-level deep auto-encoder. This is a modality-level
random masking similar to the way denoising auto-encoders are trained
with random masking of the input vector elements~\citep{VincentPLarochelleH2008-small}.

At test time, any subset of modalities can be sampled given a subset of any
of the other modalities, using the same approach as described in the
previous section: compute $h^{(t)}_{L-1}=\tilde{f}^{(t)}_{L-1}(x^{(t)})$
for the observed modalities, and start a Markov chain for the top-level
auto-encoder, keeping the observed modalities clamped, and initializing the
missing modalities using the appropriate missingness masks.  After samples
of the missing modalities are obtained, project them back into their input
spaces through $\hat{x}^{(t)}=\tilde{g}^{(t)}_{L-1}(h^{(t)}_{L-1})$.

Note that using the missingness input pattern to represent the marginal
over one modality or the joint between a subset of modalities is very
efficient, but it could well be that slightly better inference (for the
reconstruction of the missing modalities) could be obtained by running a
Markov chain over the missing modalities and averaging the corresponding
reconstructions.

\section{Target Propagation for Recurrent and Recursive Networks}

The ideas introduced in the previous sections can be generalized
to handle training of recurrent and recursive networks by target propagation.
The basic idea is that the past can be seen as side information
or context for the present and future, and we want the auto-encoding structure
to model the joint of past, present and future. In turn, this allows
reconstructions of past activations based on present observations
to become targets for improving past activations.

There are two fundamental differences between the situation of 
a feedforward stack of layers that we have discussed up to now
and the recurrent network case, which can be revealed by
inspecting the unfolded graph of computation in the recurrent
case and comparing to the computational graph of the stack
of feedforward representations, and associating each
``level'' with a ``time step'':
\begin{enumerate}
\item There are visible inputs (the sensory inputs that can be clamped)
coming into {\em each level}, i.e., the representations computed at each time step.
\item The same parameters are used to map the representation at ``level'' $t$
to ``level'' $t+1$, and vice-versa.
\end{enumerate}
We can apply exactly the same recipe that we have described up to
now for a stack of feedforward representations, with these two modifications.
In the forward direction we obtain a regular recurrent net computation.
The backward direction is now associated with a {\em reconstruction network}, but it is
{\em also a recurrent network}, and it is a {\em recurrent network
without inputs}. We know that such recurrent networks without input
tend to compute either a fixed point or a Markov chain (the latter if
noise is injected). 

The learning procedure now looks like this:
\begin{itemize}
\item For the ``single-layer'' auto-encoding objective: 
 \begin{enumerate}
 \item At each time 
  step $t$, compute the representation $h_t={\rm encode}(x_t)$ of $x_t$, and
  the forward recurrent net computes a new state \mbox{$s_t=f(s_{t-1},h_{t-1})$}
from $s_{t-1}$ and input $x_{t-1}$, where $h_{t-1}$ is a representation of $x_{t-1}$
(e.g., the top level of a stack of auto-encoders, as discussed previously)
  \item The backward recurrent net makes a short-loop reconstruction $(\tilde{s}_{t-1},\tilde{h}_{t-1})=g(s_t)$.
  \item $(s_{t-1},h_{t-1})$ is used as a target for $(\tilde{s}_{t-1},\tilde{h}_{t-1})$ (to update $g$)
  \item $s_t$ is used as a target for $f(\tilde{s}_{t-1},h_{t-1})$ (to update $f$)
  \end{enumerate}
\item For the ``long-loop'' paths:
 \begin{enumerate}
 \item From {\em any} $s_T$ that has been computed by the forward recurrent net,
 run the backward recurrent net from $\hat{s}^T_T=s_T$, 
 producing at each backward step $(\hat{s}^T_t,\hat{h}^T_t)=g(\hat{s}^T_{t+1})$.
 \item $(s_t,h_t)$ is used as a target for $(\hat{s}^T_t,\hat{h}^T_t)$ (to update $g$)
 \item $\hat{s}^T_t$ is used as a target for $s_t$ (to update $f$).
 \item $\hat{h}^T_t$ is used as a target for $h_t$ (to update the encoder from $x$ to $h$).
 \end{enumerate}
\end{itemize}
A similar recipe can be applied to recursive networks, by thinking about
the associated computational graph, both forward going (up the tree)
and backwards going (down the tree).

Now this looks very much like the situation of backprop through time, where
one has to store all the past activations in order to run the long-loop
computation.  However, something interesting we may take advantage of
is that the backwards recurrent net can
be run without requiring full storage of the past sequence (it only needs
some starting points $s_T$, which could simply be the current state at the
time where the backwards procedure is run). Another interesting element
is that we really want to consider all $T$'s and not necessarily just the
last one, which could help average over many backwards-calculated targets. 
It is only for the updates themselves that it seems that we need to store the past history.
Let us consider the special case where each layer is of the
usual affine + non-linearity type found in artificial neural networks.
For long-term dependencies with $T\gg t$ the resulting $\hat{s}^T_t$
will converge to a fixed point $\mu$, and the ingredients for the resulting gradient updates
can be obtained by accumulating appropriate sums {\em forward} in time 
while at the same time estimating the fixed point $\mu$ of the backward network.
So this addresses both very short dependencies (the short loop) and
very long ones (the fixed point), but the question of capturing intermediate
ones without having to store the corresponding 
sequences remains open (and maybe there is no other way but storing long sequences).

In both the case of the recurrent net and of the recursive net, we can
think of the architecture as a very deep tree-structured auto-encoder with
shared weights, the only difference being that in the case of a recurrent
net, the tree is unbalanced and is really a chain with dangling
leaves. Note how this deep auto-encoding structure allows one to resample any part of
the sequence given any part (by doing stochastic encode/decode steps in
which only the missing elements are resampled).

What is interesting is that we have potentially removed backprop from
the picture of training a recurrent or recursive network. It would
be very interesting to see if target propagation allows to train
recurrent networks to capture longer-term dependencies than backprop-based
training.

\section{Making the Auto-Encoders Perfect}

The initial discussion on the layer-wise $KL$ training criterion
and the use of deterministic encoders and decoders promoted the objective
that the encoder/decoder pairs should be trained to be near inverses
of each other for inputs that come respectively
from $Q$ (for the encoder) or from $P$ (for the decoder). 
Is that a reasonable objective? Note that we do not mean that
the auto-encoders necessarily invert {\em any} $x$ and {\em any} $h$.
Only that they do it almost perfectly for {\em almost any sample}
from respectively $Q$ or $P$. If $Q(X)$ lives near a low-dimensional
manifold, then the encoder can throw away unnecessary dimensions
and thus not be invertible for unlikely input configurations.

As argued above, one motivation for considering the extreme case of perfect auto-encoders
rather than assuming some stochastic reconstruction distribution $P(X|h)$
and stochastic encoder $Q(H|x)$ is that if we parametrize the auto-encoders
in a non-deterministic way, we may end up with noise added in the generative
process for $P(X)$, yielding much more noisy samples than the training data.

Another, more fundamental motivation for considering perfect auto-encoders 
is that getting an auto-encoder pair to be almost perfect is not really
difficult if the code dimension is sufficient, 
and that what is difficult instead is to make the encoder
transform a complicated distribution ($Q(X)$) into a simple one ($Q(H)$),
in the sense of being easier to model (which intuitively means ``flatter''
or more easily factorizable).
The proposal of this paper is to make this transformation gradual, 
with each layer contributing a little in it.

Perfect reconstruction on the data points
can be achieved automatically in various ways (maybe not
all desirable). For example, if the encoder is an optimization procedure
that looks for
\begin{equation}
  \tilde{f}(x) \in {\rm argmax}_{h \in S} \log P(x | h)
\end{equation}
then we get perfect reconstruction so long as $S$ is large enough
to have a separate value for each $x$ in the training set.
For example, if $x \sim Q(X)$ lives on a $d$-dimensional manifold,
then $S=\mathbf{R}^d$ could be sufficient to get perfect reconstruction.

Another interesting example, that is computationally less demanding,
is to make each layer of the encoder easily invertible. For example,
consider the usual non-linear transformation in neural networks,
with $f_l(h_{l-1}) = \tanh(b_l + W_l h_{l-1})$, where $h_l$ is the
$l$-th layer representation vector and $W_l$ is a matrix of weights
and $b_l$ a vector of biases. The hyperbolic tangent
is invertible, and if we make $W_l$ an invertible square $d \times d$ matrix, then
we can in principle compute the inverse for cost $O(d^3)$. If we
choose minibatches of length greater then $d$, then inverting
the weight matrix is of the same order as computing the matrix
multiplication. Even
better, we might be able to parametrize $W_l$ so that it is invertible
for a cost $O(d^2)$, which is the same as the matrix-vector product.
For example if $W_l = L L'$, the product of a lower-diagonal matrix
and its transpose, then the inverse of $W_l$ can be computed by
forward and backward-substitution in $O(d^2)$.  Another interesting
possibility is to decompose $W_l = U D V'$ where $U$ and $V$
are maintained nearly orthogonal and $D$ is diagonal. Maintaining
exact orthogonality can be expensive, but maintaining approximate
orthogonality is easy ($U$ and $V$ just need to be the encoders of linear
auto-encoders with squared reconstruction error).

In general, one would expect that one can {\em learn} encoder/decoder
pairs that are near inverses of each other, and we can make that
almost a hard constraint because there are many ways in which
this can be done, leaving enough degrees of freedom for other
desiderata on the auto-encoder, such as making the distribution simpler as we move
up the ladder of representations.

Note that in order for the encoder/decoder pairs to be perfectly matched
for the distributions that they see, it is important that the layers have
sufficient size. If the dimension of $h_i$ is too small relative to the
dimension of $h_{i-1}$, then the decoder will not be able to do a nearly
perfect job on the training data. The dimension of $h_i$ can be reduced
with respect to that of $h_{i-1}$ only to the extent that the data really
lives in a lower-dimensional manifold.  Even then, the reduction should be
gradual because the work of compression may be best done in a series of
gradual non-linear steps. What we recommend is to actually {\em keep all
  the layers of the same size}, but use means other than the layer size to
obtain a compression and a contraction. We know that the denoising
criterion automatically yields a contraction~\citep{Alain+Bengio-ICLR2013},
so we do not need to impose an explicit one, although it might be
interesting to experiment with alternative ways to encourage contraction,
such as sparsity or the contraction penalty of contractive
auto-encoders~\citep{Rifai+al-2011-small}.

If we keep all the latent layers of roughly the same size, then we might
want to have either the encoder or the decoder equipped with an
intermediate hidden layer that is not considered to be part of
the set of latent layers $h_i$. Instead, such an intermediate hidden layer would simply
be considered as part of the computation for the layer-wise encoder
or decoder, e.g., the encoder is an affine+rectifier transformation
and the decoder is an MLP, or vice-versa. We also have indications that
if one fixes the decoder (e.g., to some parametric transformation), then
the generally optimal encoder is non-parametric (and can be obtained
via an optimization, like in sparse coding). Since an optimization
is computationally expensive, we could replace it by a high-capacity MLP.
If the encoder or the decoder is an MLP, then we still have to use
back-prop internally to train it, but we know that training a shallow
neural network by back-prop with gradient-based optimization works well,
so this is not a big concern.

Then the question is whether it should be the encoder or the decoder that
is equipped with a higher capacity (and internal hidden layer), or
both. This question should be determined experimentally, but a practical
concern is that we would like the recognition path to be fast (and it would
be good for brains as well), in order to be able to make fast inference and quick
decisions. That suggests that the encoder should be a ``simple'' non-linear
transformation (like the usual neural network layer) while the decoder
should be an MLP, but this should be resolved experimentally.

\section{About the Top-Level Auto-Encoder and Avoiding a Top-Level MCMC to Sample}

\subsection{Linear Top-Level Auto-Encoder}
\label{sec:linear-top}

In the special case where the top-level is linear, training it to minimize
denoising reconstruction error estimates a multivariate Gaussian prior for
$P(H_{L-1})$. One can readily verify that the optimal denoising
reconstruction function minus the input $h_{L-1}$ behaves similarly to the gradient of the
log-likelihood, $\Sigma^{-1} (h_{L-1} - \mu)$, which pushes back towards
the mean in a stronger way in directions of smaller eigenvalue.
However, note how the true gradient blows up if some eigenvalues are 0,
unless $h_{L-1}$ happens to be already lying on the allowed manifold. But
even in that case, we get a numerically unstable result, dividing a 0 (the projection
on the 0-eigenvalue direction) by a 0 (the eigenvalue with value 0).
Any slight perturbation of $h_{L-1}$ would throw this off. On the other hand,
one would clearly get a stable reconstruction if the system is
trained as a linear denoising auto-encoder, because it always sees
a numerically bounded reconstruction target, and is trained with
stochastic variations of $h_{L-1}$ in the first place.

Viewing the top-level auto-encoder as a Gaussian however has the
advantage that one can replace the MCMC sampling scheme of general denoising
auto-encoders by the analytic sampling scheme of a Gaussian.
The auto-encoder weights can be mapped to the Gaussian covariance
(and the biases to the mean) by a simple calculation.
However, it is unlikely that every input distribution can be mapped to
a Gaussian distribution, simply because we know that there are 
discrete factors typically involved (e.g., multiple disjoint manifolds).
What really makes the Gaussian easy to sample, though, is that it
affords a completely factorized representation, where the factors
are statistically independent of each other.

\subsection{Factorial Top-Level Prior Instead?}
\label{sec:factorial-top}

Hence if we can map to a factorized top-level distribution (possibly
with both discrete and continuous variables), then we can generate
$(x,h)$ through completely ancestral sampling, where each step is
exact, rather than having to rely on an MCMC for the top level.
One interesting question is whether every ``reasonable'' distribution
can be mapped through a generally non-linear but invertible transformation
into a completely factorized one (and what ``reasonable'' then entails).

Note that in an ordinary stack of auto-encoders, 
the top-level auto-encoder does not have a prior that regularizes
its code layer, unlike the lower auto-encoder layers. This is equivalent
to saying that the top-level prior has a zero gradient, meaning that
it has a constant probability, i.e., a completely flat probability
distribution, such as the uniform or a large variance Gaussian.
Forcing a top-level uniform distribution may be too strong, and
it seems that a weaker assumption is that the top-level prior
is simply factorial. As discussed above, that makes generative sampling
very easy and also makes it more likely that the top-level factors
have some intrinsically interesting meaning that can be
revealed through visualizations. The idea would thus be that
the top-level prior is not really an auto-encoder, or is a ``diagonal'' one
that reconstructs every unit $h_{L,i}$ separately given itself.

If we choose the top level to be an arbitrary factorial distribution, then
instead of doing a reconstruction in order to estimate $\frac{\partial \log
  P(h_L)}{\partial h_L}$, we can just compute analytically this derivative,
for continuous hidden units. What should the equivalent target
be for discrete top-level variables? 
A plausible answer is that the target reconstruction for a discrete unit taking
values in some set $S$ should simply be the mode of that discrete
distribution. If the unit is binary, it just means that the reconstruction
is either equal to the input or to its complement, whichever is most 
probable. One worrisome aspect of this is that when we do this on every
bit of the top level discrete units, we get a target reconstruction
that may be very far from the actual output. Another option is to
consider a reconstruction which is continuous even though it is
a reconstruction of discrete variables. This is typically what we
have with auto-encoders trained with discrete inputs: the reconstruction
units compute the probability of these values. In the case of
a factorial distribution, the ``reconstruction target'' is thus
just the prior probability itself.

\subsection{Parzen Top-Level Instead}

Another interesting possibility is to make the top level of the
hierarchy a Parzen distribution, with a regularizer that pushes
the variances of each component to be as large as possible.
The proposal here ties in well with the idea of training
with a nearest-neighbor reconstruction criterion introduced
in Section~\ref{sec:nearest-neighbor}.

The ``reconstruction'' of an example $x$, seen as $h_L$, is
then basically a linear combination of the nearest neighbors,
weighted by their relative component probability. More precisely,
\begin{align}
P(h_L) \propto& \sum_i\; e^{-\frac{1}{2}\frac{||h_L - \mu_i||^2}{\sigma^2}} \nonumber \\
w_i =& \frac{e^{-\frac{1}{2}\frac{||h_L - \mu_i||^2}{\sigma^2}}}
            {\sum_j e^{-\frac{1}{2}\frac{||h_L - \mu_j||^2}{\sigma^2}}} \nonumber \\
\frac{\partial \log P(h_L)}{\partial h_L} =& \sum_i w_i \frac{(\mu_i - h_L)}{\sigma^2}
\end{align}
where $\mu_i$ are the Gaussian means, i.e., a set of training examples excluding $h_L$,
and $\sigma$ is the bandwidth of the Parzen windows.

This should push $h_L$ towards the nearest mode as estimated by the Parzen
density estimator, i.e., the nearest neighbor or the convex set spanned
by a few nearest neighbors. It thus has the expected contraction property,
but unlike the previous models, it allows a very high-capacity top level.

Ancestral sampling from the top-level is of course very simple: randomly choose one
of the stored templates $\mu_i$ (i.e., the representation of some
actual example), add Gaussian noise with variance $\sigma$, and
project the result back through the decoder, into input space.

In order to avoid adding noise in unwanted directions that would
yield poor generated samples, two possibilities arise, which
can be combined:
\begin{itemize}
\item Make the Parzen model more sophisticated, making it
more concentrated in some directions, e.g., using a local covariance
estimator at each location, as with Manifold Parzen Windows~\citep{Bengio-Larochelle-NLMP-NIPS-2006-short}.
\item Make the decoder contractive in the appropriate directions. This
will happen if $Q(h_L|x)$ adds noise in the same isotropic way 
as $P(h_l)$ is, i.e., by adding Gaussian noise of variance $\sigma^2$
to the encoder output. The samples from $Q(h_L|x)$ will be the input
to the decoder, and the decoder will have to learn to be insensitive
the added noise. The pro-entropy term of the $KL$ criterion (Eq.~\ref{eq:KL})
helps to calibrate the amount of noise (and potentially the directions of it)
appropriately in order to balance reconstruction error.
\end{itemize}

One worry with such a non-parametric top level is that it only allows
local generalization in the space of high-level factors. It will not generalize
``far'' from the training examples in a combinatorial sense like a factored
model or a denoising auto-encoder could, i.e., to
obtain an exponentially large set of configurations of the top-level factors.

\subsection{Learning How to Corrupt at the Top Level}
\label{sec:learn-corruption}

An interesting question is how to choose the amount of corruption
to be injected at the top encoder, i.e., when sampling from $Q(H_L|h_{L-1})$.
For factorial Gaussian units, we can find answers already
in~\citet{Kingma+Welling-ICLR2014}. The gradient with respect
to the prior and entropy terms in $KL(Q||P)$, i.e., all but the reconstruction term,
can be integrated out in the case where both $Q(H_L|h_{L-1})$ and
and $P(H_L)$ are diagonal Gaussians. In that case, we might as well
choose $P(H_L)$ to have the identity matrix as covariance and only
learn the variances (possibly conditional) of $Q(H_L|h_{L-1})$.
Let the $i$-th element of $Q(H_L|h_{L-1})$ have mean $\mu_j$
and variance $\sigma_j$. Then the training cost reduces to
\begin{equation}
\label{eq:integrated-cost}
  - \log P(x | h_L) - \frac{1}{2} \sum_j (\log \sigma_j^2 + 1 - \mu_j^2 - \sigma_j^2)
\end{equation}
where $h_L$ is the sample from the $(\mu,\sigma)$ Gaussian.
The first term is the usual reconstruction error term (trying to make $\sigma$ small 
and the overall auto-encoder doing a good job), 
the conditional
entropy term corresponds to $\log \sigma_j^2$ (trying to make $\sigma$ large), 
and the prior term to $1 - \mu_j^2 - \sigma_j^2$ (trying to make $\sigma$ and $\mu$ small).
To make the notation uniform in Algorithm~\ref{alg:generative-training},
we denoted $h_L$ for the mean $\mu$ and $\hat{h}_L$ for the
corrupted sample from $Q(H_L|h_{L-1})$. 

Minimizing the above integrated criterion by gradient descent on $\sigma$ allows
the model to select a proper level of injected noise for each dimension. Should $\sigma$
be a function of $x$? Keep in mind that the decoder should be able to contract
out the noise in order to do a good reconstruction job (and avoid that $P(x)$ be
too blurred because $P(x|h)$ would be blurred).  The decoder does not know in
which directions more noise has been added, and if it were to contract in all
dimensions it would necessarily contract too much (bringing every $h_L$ to
roughly the same $x$).  This suggests that the noise should be added in a
consistent way (i.e., independent of $x$), so that the decoder has a chance to
know in which directions it needs to contract most. We can think of this decoder
contraction as a form of latent variable selection: the decoder picks which
dimensions in latent space correspond to noise (and how much noise) and which
dimensions correspond to signal. We should thus be able to read the
``dimensionality'' of $x$ discovered by the model by looking at the
``spectrum'' of values of $\sigma_j$.

In the above discussion, we have only considered the case of continuous
latent variables, but as argued earlier in this paper, it is important
to have discrete latent variables as well in order to represent the
discrete aspects of the data, e.g., categories. This could be obtained
with discrete stochastic units, such as the classical stochastic
binary neurons proposed for Boltzmann machines and Helmholtz machines.
In that case, we can think of the deterministic part of the encoder
output as the sign of the pre-sigmoid activation, and it is the
magnitude of the pre-sigmoid activation which indicates the amount
of uncertainty (most uncertainty at 0) and carries the entropy.
For factorized stochastic binary units with output sigmoid probability $p_j$
we can compute the conditional entropy of $Q(H_L|h_{L-1})$ and the
contribution of $\log P(H_L)$, which together correspond to the
cross-entropy of $Q(H_L|h_{L-1})$ and $P(H_L)$. For Bernoulli units
with encoder output probability $q_j$ and prior probability $p_j$
the contribution is therefore
\[
  - \sum_j q_j \log \frac{q_j}{p_j} + (1 - q_j)\log \frac{(1-q_j)}{(1-p_j)},
\]
which can easily be differentiated with respect to the output weights
of the top-level encoder. Assuming a flat prior like in the Gaussian case,
$q_j=0.5$ and
this amounts to simply pushing the weights towards 0. However, this is specialized form
of weight decay that ``saturates'' when the weights become large, and which
depends on the global behavior of each unit rather than on individual weights only,
two properties that differ from
L1 and L2 weight decay.

A nagging question is whether there should even be any noise in the signal
directions, i.e., in the directions (i.e. dimensions of $h_L$) along which there
are variations in $x$-space in the data distribution.  Adding noise along a
signal direction means that either $P(x|h_L)$ will be blurred or that the output
conditional distribution associated with $P(x|h_L)$ will have to have a very
complex shape (which we said we wanted to avoid, leaving that job to the deep
net itself). It seems that the only way to avoid these unsatisfactory outcomes
is to only allow noise to be added in the non-signal directions of $h_L$,
which the decoder can then trivially ignore. If that hypothesis is correct,
then the only thing that needs to be learned regarding the entropic 
components of $Q(H_L | h_{L-1})$ is the following ``binary'' question:
{\em is this a component of noise or a component of signal}?
If it is signal, then no noise should be added, and if it is a component
of noise, then it is not even necessary to include that component as
an input to the decoder, thereby automatically eliminated potential problems
with $P(H_l)$ generating variations along these directions.
What is difficult with this interpretation is that we are faced with
a binary choice whose answer depends on the final state of training.
What we would like is a soft way to gradually eliminate the noise
directions from the picture as training progresses. Another option
is to do like in PCA and treat the number of signal directions as
a hyper-parameter by fixing the number of top-level signal directions
(in which case no explicit computation is necessary for modeling
or contracting the ``noise'' directions).

Note that all of these questions go away if we make the top-level
model powerful enough to match $Q(H_{L-1})$. On the other hand
a completely non-parametric top-level does not generalize in a satisfactory
way. This leaves the auto-encoder based models, which may have enough
power to match $Q(H_{L-1})$ while not requiring to explicitly
choose the number of signal dimensions.

\subsection{MAP vs MCMC for Missing Modalities and Structured Outputs}
\label{sec:MAP}

An interesting observation is that the noise-free target propagation 
provides an easy way to perform approximate MAP inference in the case of missing
modalities or structured outputs. Indeed, if we take $\hat{h}_{L-1}-h_{L-1}$
as pointing in the direction of the gradient of $\log P(h_{L-1})$
(or towards a more probable configuration, in the discrete case),
then local ascent for 
MAP inference can be achieved as outlined in Algorithm~\ref{alg:MAP-inference},
by iteratively encoding and decoding at the level of $h_{L-1}$ 
with the deep auto-encoder sitting on top of $h_{L-1}$. This will
correspond to a {\em local ascent} to estimate the MAP, in the
space of $h_{L-1}$. Iteratively updating the input by deterministic 
encode/decode steps of a trained denoising auto-encoder was successful
done by~\citet{Bahdanau+Jaeger-2014} in order to find a local mode (nearby MAP estimate),
around the starting point of the iterations. They used the {\em walk-back} variant of the training
criterion~\citep{Bengio-et-al-NIPS2013-small} in order to make sure
that most spurious modes are removed during training: otherwise
the iterated encode/decode step tend to go to these spurious modes.
It would probably be a good idea to use a walk-back variant here: it
simply means that a target (the uncorrupted input of auto-encoder)
is provided for the reconstruction after multiple noisy encode-decode steps,
and not just one.

\begin{algorithm}[ht]
\caption{MAP inference over some subset of a representation  $h_{L-1}^{\rm (missing)}$ (e.g. associated
with a structured output target $y$, or some missing modalities),
given the rest,  $h_{L-1}^{\rm (observed)}$ (e.g., associated with an input $x$, or some observed
modalities).}
\label{alg:MAP-inference}
\begin{algorithmic}
\STATE Compute $h_{L-1}^{\rm (observed)}$ for the observed parts of the data,
through their respective encoder functions.
\STATE Initialize $h_{L-1}^{\rm (missing)}$ (e.g. to some mean value, preferably set to be 0 by construction).
\STATE Set corresponding missingness masks as inputs (only for the first iteration below, then consider all
inputs into the top-level auto-encoder as observed (having been filled-in iteratively).
\REPEAT
  \STATE Let $\hat{h}_{L-1}$ be the reconstruction of the top-level auto-encoder (with no noise injected)
  taking $h_{L-1}=(h_{L-1}^{\rm (observed)}, h_{L-1}^{\rm (missing)})$ as input.
  \STATE Update $h_{l-1}^{\rm (missing)}= \hat{h}_{L-1}^{\rm (missing)}$ while keeping the observed parts fixed.
\UNTIL a maximum number of iterations or convergence of $h_{L-1}^{\rm (missing)}$
\STATE Map $\hat{h}^{\rm (missing)}_{L-1}$ deterministically back to data space through associated decoder functions.
\STATE {\bf Return} the resulting predicted missing values $\hat{x}^{\rm (missing)}$.
\end{algorithmic}
\end{algorithm}

In general,
we are not interested in finding the global MAP configuration, 
\[
 {\rm argmax}_{h_{L-1}} \log P(h_{L-1})
\]
but rather a conditional
MAP, e.g., if we want to predict the MAP output given some input,
or if modalities are observed while others are missing
and we want to infer a probable value of the missing ones:
\begin{equation}
  {\rm argmax}_{h_{L-1}^{({\rm missing})}} \, \log P(h^{({\rm missing})}_{L-1} | h^{({\rm observed})}_{L-1}).
\end{equation}
This can be achieved simply by clamping the $h^{({\rm observed})}_{L-1}$ to their
observed values and only update the $h^{({\rm missing})}_{L-1}$ in the 
deterministic encode/decode iterations. As discussed above, the target $\hat{h}$ 
(minus $h$) associated with
a subset of the elements of $h$ can be interpreted as a gradient direction, so
changing only those while keeping the others fixed amounts to a form
of gradient ascent for the missing components given the observed components.

Interestingly, the MCMC version is structurally identical, except that
noise is injected in the process. By controlling the amount of noise,
we actually interpolate between a MAP-like inference and an MCMC-like
posterior sampling inference. Note that in many applications, we
care more about MAP inference, since a specific decision has to be taken.
However, starting with an MCMC-like inference and gradually reducing
the noise would give rise to a form of annealed optimization, more likely
to avoid poor local maxima of the conditional probability.

\section{``I think I know how the brain works!''}

The title of this section is a quote often attributed to Geoffrey Hinton. 
The desire to understand, in particular, how the brain {\em learns}, has motivated
many of his very influential ideas, ideas and motivations which in turn
have motivated this work and its author.

How could the algorithmic and mathematical ideas presented here be turned
into a biologically plausible mechanism for learning deep representations
of the kind that have been so successful in recent
industrial applications of deep learning, such as speech and object
recognition? Could it even provide better representations in the case of
very deep networks?

Neurons presumably use essentially the same mechanism at every time
instant, and every (fast) signal that is used to drive learning must be
obtained locally from the output of other neurons.  This kind of constraint
has made the idea of biologically plausible implementations of
back-propagation somewhat difficult, although something clear come out of
previous attempts~\citep{Xie+Seung-2003,Hinton-DL2007}: it is likely that
feedback connections, especially those coming from downstream (moving away
from the sensory neurons) areas, are involved in providing some kind of
training signal for the upstream computations (arriving on the paths from
the sensory areas).  The proposal discussed here is in particular close to
the idea introduced by~\citet{Hinton-DL2007} that feedback connections into
a neuron modify it slightly towards values that would reduce some ulterior
loss function, i.e., that gradients are implemented in the brain as
temporal derivatives. That talk also made the point that if the feedback
connections implement a good auto-encoder and the weights are symmetric,
then the CD-1 update will be close to training both the encoder and the
decoder to minimize reconstruction error.  Consider activations vector
$h^i=\sigma(a^i)$, where $\sigma$ is a differentiable activation function
and $a^i=W_i h^{i-1}+b_i$, along with a reconstruction path with the
transpose weights, $\alpha W_i^T$ for $\alpha$ small. \citet{Hinton-DL2007}
justified the proposed biological implementation of back-propagation on the
grounds that if the temporal derivatives $\dot{h}^i$ of activations
represents $\frac{\partial C}{\partial a^i}$, then the pertubation of $h^i$
translates into a perturbation of $h^{i-1}$ which is proportional to the
derivatives through the feedback reconstruction path, including the
derivative of $\sigma(a^{i-1})$ with respect to $a^{i-1}$, making
$\dot{h}^{i-1}$ proportional to $\frac{\partial C}{\partial a^{i-1}}$.
Here we exploit another kind of justification for the use of feedback paths
to perform credit assignment: the relationship between reconstruction and
score (gradient of log-likelihood) that has been discovered in recent
years~\citep{Vincent-2011,Swersky-ICML2011,Alain+Bengio-ICLR2013}.
The advantage of the latter mathematical framework is that it does not rely on
symmetric weights nor on the specifics of the parametrization of the
encoder and decoder, only on the requirement that feedback paths
are trained to reconstruct, i.e., so that the encoder and decoder fibers
try to invert each other for observed inputs and contract other configurations
towards observed ones, thus making the encoder-decoder loops
map neuron configurations to more likely ones according to the learned
model of the world.

The training criterion introduced in this paper suggests that there
are two main elements that should drive synaptic updates:
\begin{enumerate}
\item {\em Prediction}: the incoming synapses (especially on feedforward
paths, if such a thing is well-defined in the brain) should change so as
to make the neuron activation closer to what it will be a bit later,
after the rest of the network (possibly over multiple loops and corresponding time
scales) has pushed the activation towards values that are closer to
what the overall network ``likes to see'' (more probable configurations
according to the implicit model learned). There is already plenty
of evidence of such a synaptic change mechanism in the form of
the spike-timing dependent plasticity (STDP).
\item {\em Reconstruction}: the incoming synapses (especially on
feedback paths, if such a thing is well-defined int the brain) should
change such that every loop forms a better regularized auto-encoder.
One way to achieve this would be for these synaptic synapses to change
so as to make the neuron activation (now viewed as a reconstruction)
closer to what it was a bit earlier (viewed as the original value before
getting the feedback from the rest of the network). Another way
to achieve this would be to also use the forward-prediction update
(i.e. STDP) but to arrange the timing of things so that an initial
noisy computation (e.g. a volley of spikes quickly travelling through
the brain) is followed by a less noisy computation (e.g. neurons
have integrated more spikes and compute a more reliable activation),
which serve as ``target'' for the previous noisy ``reconstructions''
(from feedback paths).
There is already some evidence
that such a mechanism might exist in brains~\citep{Hanuschkin-et-al-2013,Giret-et-al-2014}, but 
more experimental neuroscience is clearly needed here.
\end{enumerate}

It is not clear if the same neuron can play the role of propagating
signals forward (prediction: match the future) and backwards 
(reconstruction: match the past), and if so, it is not clear whether different
synapses should be explicitly assigned to the prediction and others
to the reconstruction. One may need a group of actual biological neurons
working together (e.g. in a micro-column) in order to implement the
predictive and reconstructive components discussed in this paper.
The evidence from our knowledge of cortical microstructure and layered
structure of the cerebral cortex~\citep{daCosta+Martin-2010} suggests 
that each cortical area has separate ``feedforward inputs'' and ``feedback inputs''
arriving at or going to different layers, in the sense that the connections from area A
to area B involve different layers (e.g. from L2/3 in A to L4 in B) then
the connections from area B to area A (e.g. from L2/3 and L6 in B to L2/3 in A).

Note that the only thing that prevents the above principles from collapsing
all neurons into a configuration where they are all 
outputting a constant (say 0), is the presence of ``boundary
conditions'' that are occasionally present and ``clamp'' some of the
neurons due to external stimuli. For the reconstructive fibers into those clamped
neurons, the weights must change in order to try to match those
clamped values. In turn, when the stimuli are not present, 
there is a feedback loop through those ``visible units'' back
up into the intermediate layers, and by the ``reconstruction principle''
these up-going connections into a neuron $n$ must learn to invert the down-going
connections from $n$ to an (unclamped) sensory neuron. This forces the
first ``layer'' to be an auto-encoder, and now there is pressure on the
neurons further downstream to encode the output of that first layer.
As the neurons further downstream start to improve at modeling their input,
their down-going reconstruction connections now provide an additional signal (through the
predictive principle) to the first layer, telling it not just to
capture the information in the sensors, but also to transform it in such
a way as to make it easier for the upper layers to model the first layer.
Similarly, the rich statistical dependencies present in the input 
rise to higher levels of the neural circuit.

Conveniently, when the input is a temporal stream, both of these principles
coincide to some extent (if the forward-going neuron and the backward-going
neuron are the same) with the temporal coherence
prior~\citep{becker+hinton:1993,wiskott:2002,hurri+hyvarinen:2003-small,kording2004-small,cadieu+olshausen:2009-small},
sometimes known as ``slowness'', and found in slow feature analysis. It
encourages the learned features to capture abstractions which do not change
quickly in the input stream, and high-level abstractions tend to have that
property. Since different concepts change at different time scales, one
could imagine variants of the above principles where the time scale $\Delta$
is different in different neurons, corresponding to different levels in a
temporal hierarchy.

One important element to consider in any follow-up of this proposal
is that signals will travel up and down through all the possible loops,
{\em returning to a neuron $n$ after different delays}. We already
see such a scenario in the experiments on deep Generative Stochastic
Networks, see Figure 2 (right) of ~\citet{bengio-et-al-ICML2014-small},
and reconstruction error at each time step (corresponding to loops
of different lengths through the network) and the corresponding
gradients are simply added.
Hence multiple
$\Delta$s must be considered in the application of the above principles.
The smallest $\Delta$ corresponds to the shortest loop (e.g. going one
layer above and back down) while the largest $\Delta$ corresponds
to going to the other side of the brain and back. Fortunately, one
{\em does not need to store the past activations} for that range
of $\Delta$ values. This is because the update rule for synaptic
weights is additive and linear in the various targets associated with different
$\Delta$. 

Consider the
simple case of binary sigmoidal neurons with affine activation. The
delta rule for a synapse with a weight $w_i$ associated with a pre-synaptic 
input $x_i$ and an output activation \mbox{$p={\rm sigmoid}(b+w\cdot x)$} with target $t$
is 
\[
  \Delta w_i \propto (t - p) x_i. 
\]
The update on a ``feedforward'' connection $w^f_i$ ($f$ for forward or future,
with post-synaptic activation $p^f$)
should then be proportional the current post-synaptic value $p^r$ of the reconstruction
neuron (now considered a target)
times the moving average
of past pre-synaptic $x_i$ associated with $w^f_i$, minus the average
past values of $p^r x_i$: 
\[
  \Delta w^f_i \propto (p^r \overline{x_i} - \overline{p^f x_i})
\]
where bars on top indicate the moving average of past values.

The update on a ``feedback connection'' $w^r_i$
would be proportional to the current associated pre-synaptic value $x_i$ times the 
difference between the average of the past
post-synaptic forward values, $\overline{p^f}$ and the current reconstruction $p^r$:
\[
  \Delta w^r_i \propto (\overline{p^f} - p^r) x_i.
\]

\section{Conclusions, Questions, Conjectures and Tests}

In this paper we have proposed a radically different way of training
deep networks in which credit assignment is mostly performed thanks
to auto-encoders that provide and propagate targets through the
reconstructions they compute. It could provide a biologically
plausible alternative to backprop, while possibly avoiding some of
backprop's pitfalls when dealing with very deep and non-linear
or even discrete non-differentiable computations.

This approach derives primarily from the observation that regularized
auto-encoders provide in their reconstruction a value near their input
that is also more probable under the implicit probability model that
they learn, something that generalizes gradients, and can thus be
used to propagate credit (targets) throughout a neural network.

This approach has first been derived from a training criterion that
tries to match the joint of data and latent representations when
they are generated upward (from the data generating distribution and 
upward through the encoder) or downward (from the learned generative
model, and going down through the decoders). This criterion
is equivalent to the variational bound on likelihood that has 
previously been used for graphical models with latent variables.

This paper discusses how this idea can be applied to a wide variety
of situations and architectures, from purely unsupervised and generative
modeling to supervised, semi-supervised, multi-modal, structured 
output and sequential modeling.

However, many questions remain unanswered, and the main ones are
briefly reminded below.

\begin{enumerate}
\item {\bf Do the Level-Wise Targets Help?}

In Algorithm~\ref{alg:generative-training} or \ref{alg:supervised-training},
we have targets at each layer and we could back-propagate the associated
targets all the way or we could only use them for updating the parameters
associated with the corresponding layer. Do the intermediate targets make
backprop more reliable, compared to using {\em only backprop}?
(in the generative case the comparison point would be the variational
auto-encoder with no intermediate latent layer, 
while in the supervised case it would be the ordinary
supervised deep net).

\item {\bf Can Back-Propagation Between Levels be Avoided Altogether?}

Following up on the previous question, can we use {\em only} the
propagated targets and no additional backprop at all (between layers)?
We might still want to use backprop inside a layer if the layer
really is an MLP with an intermediate layer.

\item {\bf Can we Prove that a Denoising Auto-Encoder on Discrete Inputs 
Estimates a Reconstruction Delta that is Analogous to a Gradient?}

We already know that reconstruction estimates the log-likelihood gradient
of the input for denoising auto-encoders with small noise and continuous inputs.
What about larger noise? Is there an analogous notion for discrete inputs?
A related question is the following, which would be useful to answer if
we want to have a factorial top level.
{\bf What would be an appropriate reconstruction target in the case of a factorial
discrete distribution?}

\item {\bf Are we Better Off with a Factorial Top-Level, a Gaussian Top-Level or with an Auto-Encoder
Top-Level?}

Related to the previous question, how should we parametrize the top level? An explicit
top-level factorized distribution is advantageous because one can sample analytically
from it. A linear auto-encoder is equivalent to a Gaussian, and can also be sampled
analytically, but cannot capture discrete latent factors.

\item {\bf Can we Prove that Algorithm~\ref{alg:generative-training} is
a Consistent Estimator, with Enough Levels?}

If we provide enough levels, each with dimension of the same order as the input,
and if we train each level according to Algorithm~\ref{alg:generative-training},
are there conditions which allow to recover the data generating distribution
in some form? 

\item {\bf Can Every Reasonable Distribution be Mapped to a Factorial One?}
Is there a transformation $f(x)$, not necessarily continuous, that maps $x \sim Q(X)$
to $h=f(x)$ such that the distribution of $h$'s $Q(H)$ is factorial, for any or a very
large class of data distributions $Q(X)$?

\item {\bf Is Nearest-Neighbor Reconstruction Yielding Better Models?}
Instead of reconstructing the clean input, Section~\ref{sec:nearest-neighbor}
proposes to reconstruct a near neighbor from the training set, minimizing
the discrepancy between the representations of the near neighbors
in representation space. Since that was the motivation, does that approach 
yield better mixing and more accurate generative models or better
approximate MAP, compared to ordinary reconstruction error?

\item {\bf How to Handle Ambiguous Posteriors?}
In this paper we have not addressed the question of ambiguous posteriors, i.e.,
the data is really generated from factors whose value cannot be completely
recovered from the observation $x$ itself. A natural way to handle such
ambiguity would be for the top-level encoder to be stochastic, like in the
variational auto-encoder~\citep{Kingma+Welling-ICLR2014,Rezende-et-al-arxiv2014}.
Another idea is to introduce auxiliary ``pseudo-input'' variables to the top-level
auto-encoder (in the same way we treated the labels $y$ in Section~\ref{sec:supervised}, but
being fully unobserved) that will correspond
to the real ``top-level disentangled factors of variation'' idealized
in~\citet{Bengio-Courville-Vincent-TPAMI2013}.

\item {\bf How to learn to identify noise from signal at the top signal?}
A question that is related to the above is how to determine the amount and location of
noise injected at the top level (by sampling from $Q(H_L|h_l)$). Should there
even be any noise added in the signal directions? At least at the end of training,
we should strive to have zero noise added in these directions, but we do not generally
know ahead of time how many signal directions there are, and although it could be
a hyper-parameter, this would be an unsatisfying answer.

\item {\bf Encoder or Decoder as MLP, Neither, or Both?}

Each layer of the proposed architecture does not have to be the
traditional affine layer (composed with point-wise non-linearity). It could
be more powerful, e.g., it could be an MLP with its own hidden units.
Should we use such powerful layers at all? in the encoder? in the decoder? in both?

\item {\bf Is Corruption Necessary in the Lower Layers?}

When training a deep network such as discussed here, should we limit the injection of
corruption to the upper layers only (which capture the stochastic aspect of the
distribution), while keeping the lower layers deterministic? Instead of a denoising
criterion to achieve contraction in the lower layers, an explicit contractive
or sparsity penalty could be used.

\item {\bf Is the use of a missingness input adequate to deal with reconstruction
in the presence of missing modalities or would an MCMC over the missing modalities
yield more accurate reconstructions?}

Training the top-level auto-encoder in a multi-modal system to automatically handle
missing modalities by including a missingness input makes for very efficient training
and use of the auto-encoder. The question is whether the MCMC approach (iteratively
resampling the missing modalities) would provide more accurate targets for 
training the lower levels of the observed modalities.

\item {\bf Can a Recurrent Network be Trained to Capture Longer-Term Dependencies by 
Target-Propagation then by Back-Propagation?}

Can the idea of propagating targets (obtained as reconstructions) be used to replace
backprop for training recurrent nets to capture long-term dependencies?
Can the fully storage of state sequences be avoided for intermediate length dependencies?
Since the backwards (reconstruction) recurrent net runs without input, it should
be quite possible to learn the equivalent backwards step associated to different
time scales and use it to jump over longer time horizons.

\item {\bf Does the ``learn to predict the past and the future'' principle
hold up in simulations of real-time biologically inspired neural circuits? Do associated
mechanisms exist in brains? How about temporally varying inputs?}

This paper's mathematical framework suggests a simple learning principle for
brains and it would be interesting first to verify by simulation if it allows
brain-like circuits to learn in real-time (with presumably no global
discrete clock), and second, to search for associated biological mechanisms
through physiology experiments on animals. Finally, it is important to consider
when the input is temporally varying. The proposed learning principle
amounts to training a recurrent network to predict both its future state
and its past state. The past-predicting signal in turn is used to provide
a proxy for a gradient to improve predicting the future. In order to
predict farther into the future, one could potentially use neurons that specialize
on predicting events at longer horizons (both forward and backwards in time) 
and that correspondingly
average their gradients over longer horizons, or that could be done implicitly
thanks to the long-term memory that architectures such as the echo-state
network / liquid memory recurrence~\citep{Maas-et-al-2002,Jaeger+Haas-2004} provide.

\end{enumerate}

\subsection*{Acknowledgments}

The author would like to thank Jyri Kivinen, Sherjil Ozair, Yann Dauphin, 
Aaron Courville and Pascal Vincent for their feedback,
as well as acknowledge the support of the following agencies for
research funding and computing support: NSERC, Calcul Qu\'{e}bec, Compute Canada,
the Canada Research Chairs and CIFAR.

\bibliography{strings,strings-shorter,ml,aigaion}
\bibliographystyle{natbib}

\end{document}